\documentclass[sn-mathphys]{sn-jnl}

\jyear{2021}%
\usepackage{blkarray, bigstrut}
\theoremstyle{thmstyleone}%
\theoremstyle{thmstyletwo}%

\theoremstyle{thmstylethree}%

\begin{document}

\title[Missing Value Imputation with Deep Neural Networks]{Long-Term Missing Value Imputation for Time Series Data Using Deep Neural Networks}


\author[1]{\fnm{Jangho} \sur{Park}\footnote{This work was done when Jangho Park was with Lawrence Berkeley National Laboratory.}}
\author*[2]{\fnm{Juliane} \sur{M\"uller}}\email{JulianeMueller@lbl.gov}
\author[3]{\fnm{Bhavna} \sur{Arora}}
\author[3]{\fnm{Boris} \sur{Faybishenko}}
\author[2]{\fnm{Gilberto} \sur{Pastorello}}
\author[3]{\fnm{Charuleka} \sur{Varadharajan}}
\author[4]{\fnm{Reetik} \sur{Sahu}}
\author[2]{\fnm{Deborah} \sur{Agarwal}}

\affil[1]{\orgdiv{Artificial Intelligenc Team}, \orgname{Teikametrics}, \orgaddress{\city{Boston}, \state{MA} \postcode{02210}, \country{United States}}}

\affil[2]{\orgdiv{Computational Research Division}, \orgname{Lawrence Berkeley National Laboratory}, \orgaddress{\street{1 Cyclotron Rd}, \city{Berkeley}, \state{CA} \postcode{94720}, \country{United States}}}

\affil[3]{\orgdiv{Earth and Environmental Sciences Area}, \orgname{Lawrence Berkeley National Laboratory}, \orgaddress{\street{Calvin Rd}, \city{Berkeley}, \state{CA} \postcode{94705}, \country{United States}}}

\affil[4]{\orgdiv{Water Security Research Group}, \orgname{International Institute for Applied Systems Analysis}, \orgaddress{\street{Schloßpl. 1},  \postcode{2361} \city{Laxenburg}, \country{Austria}}}


\abstract{We present an approach that uses a deep learning model, in particular, a MultiLayer Perceptron (MLP),  for estimating the missing values of a variable in multivariate time series data. We focus on filling a long continuous gap (e.g., multiple months of missing daily observations) rather than  on individual  randomly missing observations. Our proposed gap filling algorithm uses an automated method for determining the optimal MLP model architecture, thus allowing for optimal prediction performance for the given time series. We tested our approach by filling gaps of various lengths (three months to three years) in three environmental datasets with different time series characteristics, namely daily groundwater levels,   daily soil moisture, and hourly Net Ecosystem Exchange. We compared the accuracy of the gap-filled values obtained with our approach to the widely-used R-based time series gap  filling methods \textsf{ImputeTS} and \textsf{mtsdi}. The results indicate that using an MLP for filling a large gap  leads to better results, especially when the data behave nonlinearly. Thus, our approach  enables the use of datasets that have a large gap in one variable, which is common in many long-term environmental monitoring observations.}




\maketitle

\section{Introduction and Literature Review}

Time series data are recorded in diverse  application areas ranging from earth sciences to healthcare, finance, traffic, etc. Time series  often have missing values (gaps) due to, for example,  sensor failures, collection errors, or lack of resources \cite{pedro2009, yozgatligil2013comparison,kalteh2009imputation, aissia2017multivariate}. However, often complete time series datasets are required for analysis or when these data are used as inputs to numerical models. For environmental science applications, such time series gaps could coincide with crucial times where hydrological or biogeochemical fluxes or rates are high. As an example, high stream stage conditions or fluctuations in temperature can promote carbon and nutrient cycling in hyporheic zone environments \cite{gu2012riparian, arora2019evaluating}. Thus,  it becomes crucial to impute missing time series values with highly accurate  estimates.
There are two categories of time series missing value imputation methods, namely multivariate and univariate methods \cite{thi2019edtwbi, moritz2015comparison}. The first class estimates missing values by exploiting  relationships between variables. The second class  solely relies on available observations of the time series whose gaps must be filled. 
In this paper, we study a special case where a large number of consecutive  values are missing in  a single variable of a multivariate time series.

Many different methods have been used to deal with different characteristics of missing values. For example, for small gaps in univariate time series (individual missing values), the last observation before the gap can be carried forward or the next observation after the gap backward  to estimate missing values. Other approaches, which are implemented in  \textsf{ImputeTS} \cite{moritz2017imputets} provide  linear, spline interpolation, or Kalman smoothing methods. Machine learning-based algorithms  have also been studied for gap filling purposes. For example, \cite{9237768}  estimated missing values in univariate CO$_2$ concentrations and air temperature data. This approach,  however, is not able to take advantage  of multidimensional data. 

For the imputation of missing values in the multivariate case,  the $K$-nearest neighbor method has  been successfully applied for  imputing data in the field of medical science  \cite{batista2002study}. However, this and other such applications have been limited to imputation problems where values are missing only at random.  Recently, recurrent neural networks \cite{che2018recurrent, NEURIPS2018_734e6bfc} and generative adversarial networks \cite{luo2018multivariate, zhang2021missing}  have been used for imputing values in   health-care and public air quality datasets. These studies focused exclusively on the performance and accuracy of the proposed approaches for filling  individual randomly missing    values. 

MLPs have also been successfully used for imputing missing time series values  across areas of differing complexity and applications. For example,
 \cite{lingras2008evolutionary,zhong2007rationalizing} showed that time delayed deep neural network models can impute missing values in univariate hourly traffic volumes. The referenced works focus on filling relatively short gaps (up to 168 data points) and a genetic algorithm was employed for optimizing the hyperparameters of the used  gap filling methods. 
   \cite{mahabbati2020comparison} used an MLP to fill missing values in   16 meteorological drivers and three key ecosystem fluxes in the regional Australian and New Zealand flux tower network.  The authors analyzed the performance of gap filling methods by manually changing the lags (the number of  historical time steps used for making the next time step prediction). To estimate the missing flux values, they used multiple meteorological drivers such as air temperature, wind speed, soil water content, ground heat flux, and net radiation. The architecture of their MLP was determined  by using a grid search method provided in scikit-learn  \cite{sklearn}.  
 In another study, \cite{kim2020gap} focused on filling  the gaps in  methane flux measurements using multiple input variables including  	net ecosystem exchange (NEE) of CO$_2$, latent heat flux, sensible heat flux, global radiation, outgoing long wave radiation, air temperature, soil temperature, relative humidity, vapor pressure deficit, friction velocity, air pressure, and water table height. They used an MLP with two hidden layers and determined the number of nodes by using a three-fold cross‐validation method.

Unlike the referenced studies, the main contribution of our work is a method for filling a long continuous gaps (e.g., multiple continuous months of missing daily observations) in a single variable of  a multivariate time series dataset.  Another key difference between the previous studies and our research is the size of the datasets used. Our focus is on smaller datasets that have fewer variables than the cited works. Usually, these types of datasets are hard to use and tend to be discarded in analysis and modeling efforts. Our approach would enable greater use of time series datasets that have multi-year gaps in a variable, which are common in many long-term environmental monitoring observations. In contrast to prior approaches using MLPs, our proposed gap filling method optimizes the MLP model architecture and the length of the lag with an efficient method that is based on surrogate models. Our method requires fewer resources because it is automated and uses adaptive sampling rather than evolutionary or trial-and-error approaches. 

In this paper, we present a gap filling approach that is based on the concept of Time-Delayed Deep Neural Networks (TDNN) \cite{waibel1989phoneme} with the goal of   filling a long continuous gap in one variable of a multivariate time series. 
The concept of TDNN is suitable for handling  temporal time series that are a characteristic feature of environmental datasets. 
This research extends our previous work  \cite{muller2019surrogate} in which we used  deep neural networks    to predict groundwater levels for multiple years into the future   using supporting variables such as  temperature, precipitation, and river flux data. 
In the cited work, we optimized the network architectures  with an automated  hyperparameter tuning method, which we adopt and modify in the present work to enable  gap filling.  
We assume that   the target time series variable has one large gap  and that the supporting  variables that  explain the target variable are fully observed.  Our previous work showed that MLPs are successful in making accurate predictions, and therefore,  we use an MLP for gap filling in the present study.  Our research will have an impact on application areas where gap-free long-term observations are particularly useful.

The remainder of this  paper is organized as follows.  Section~\ref{sec:2} provides a brief description of  MLP models and hyperparameter optimization (HPO) used for gap filling. Section \ref{sec: exp} presents the results of our numerical study in which we compare the performance of our proposed method with \textsf{ImputeTS} \cite{moritz2017imputets} and \textsf{mtsdi} \cite{junger2018package} for filling gaps of various lengths (three months to three years) of  three different environmental datasets. 
 Finally, Section \ref{sec: con} concludes  the paper and provides future research directions.

\section{Multilayer Perceptrons and Hyperparameter Optimization for Gap Filling} \label{sec:2}
In this section, we provide brief descriptions of MLPs and the hyperparameter optimization method used to determine its architecture.

\subsection{Multilayer Perceptron (MLP)}
\label{ssec: rewviewDL}
In this paper, we use MLPs  for regression purposes to map input  to output features. An MLP is a feed-forward artificial neural network. A generic MLP consists of an input layer, at least one hidden layer, and an output layer with  
each layer  containing multiple nodes. Each node in the input and output layers corresponds to an input feature and target variable, respectively. In order to construct an input layer that encodes  temporal information, we structure each  input such that it  contains multiple time steps. We define the amount of historical information in an input as lag $l$. Thus, with  $T$ time periods, an input is defined as $\{x_{t-l},\ldots, x_{t-1}, x_t\}$, where $t=l+1,\ldots, T$.
Our goal is to predict the value at the next  time step, and thus the output layer has only a single  node, which  corresponds to the target variable $Y$ that needs to be gap-filled.  
With this architecture, the MLP  is trained by minimizing  over a set of weights and biases using a predefined loss function  such as a root mean squared error or other metric that reflects how well the MLP approximates the data. The root mean squared error is the square root of the average difference between the actual (ground truth) and predicted values. 
In the first step of this training, linear combinations of input nodes with initial weights are constructed. Each linear combination is passed to all nodes in the first hidden layer. An activation function in the nodes receives this value and returns a transformed value. The transformed values from  all nodes in the first hidden layer are given to the nodes in the second hidden layer for further transformation. This process repeats until the output layer returns the final  transformed value.  The loss function to be minimized reflects the error between the transformed output values  and the true values. By  iteratively adjusting the weights and biases associated with each node, the loss function is minimized.  We use the Rectified Linear Unit (\textsf{ReLU}) activation function, $f(\cdot)=\max(0,\cdot)$ \cite{nair2010rectified} and our training  objective function (loss) is the mean squared error (MSE). The iterative optimization of the weights and biases  is performed with the Adam optimizer \cite{kingma2014adam},  which is a type of backpropagation algorithm based on gradient descent and the chain rule \cite{rumelhart1986learning}.

Other essential hyperparameters that influence the performance of the MLP, such as  the batch size (size of batched samples in the gradient update), the number of epochs (number of iterations used in training), the dropout rate~\cite{srivastava2014dropout}, the number of hidden layers and  nodes in each layer, and the lag value $l$,  are determined by the automated optimization method described in the next section.
We assume that the  learning and decay rates of the  stochastic gradient descent method used for training the  MLP are known and fixed. For further  details about MLPs, we refer the reader to \cite{goodfellow2016deep}.
 
 \subsection{Optimization of MLP Architectures for Gap Filling} \label{sec:HPO}

In order to improve the predictive  performance of MLPs (and deep learning models in general), the hyperparameters (model architecture) should be carefully chosen \cite{feurer2019hyperparameter}. Different hyperparameter optimization (HPO) methods have been developed and used in the literature, including  random search \cite{bergstra2012random}, a combination of grid search and trial and error \cite{larochelle2007empirical},   Bayesian optimization \cite{snoek2012practical}, and genetic algorithms~\cite{xie2017genetic}.

In this work, we adapt the HPO method proposed in~\cite{muller2019surrogate},  which interprets the training of a given architecture  as an expensive black-box function evaluation of the architecture's performance.   
Thus, the goal is to find the best hyperparameters within as few trials as possible by employing computationally efficient methods. To this end, surrogate models (in particular radial basis functions and Gaussian process models) are used to map the hyperparameters to their respective performance. An initial experimental design with $n_0$ different hyperparameter sets is created for which the performance  is evaluated by training the respective models. This initial design is used to create the computationally cheap surrogate models.  In each iteration of the optimization, an auxiliary optimization problem is solved on the computationally fast to evaluate  surrogate model  to determine the next hyperparameters to be tried.   The surrogate model is updated each time a new hyperparameter set has been evaluated. 

In order to adapt the  HPO algorithm described in   \cite{muller2019surrogate} for our gap filling purposes, we only have to change the way we split the training and validation datasets  by taking into account the fact that some data are missing. Since we include the lag (the number of historical time steps for predicting the value at the next time step) in our list of  hyperparameters to be optimized, we have to create a lagged table of the data for each set of hyperparameters we want to evaluate. 

We denote the observation matrix with  $N$ variables and time length $S$ as $ X = (x_1, x_2, \ldots, x_S)^T \in \mathbb{R}^{S\times N}$, where $x_s = (x_s^1,x_s^2,\ldots, x_s^N) \in \mathbb{R}^N$ is an observation vector at time $s, 1\leq s\leq S$. Each $x_s^d$ is an observation of the $d^{th}$ variable, where $d=1,\ldots, N$. We assume that the first column in the matrix $X$ corresponds to the target vector $Y$ that  contains missing values, and the remaining  columns are the supporting variables. In the following, we denote $x^d_s = \text{NaN}$ if the value is missing. For example, assume we have nine observations ($S=9$), one target variable, and two supporting variables  ($N=3$), then the observation matrix and target value vector are, respectively, 
\[X=
\begin{blockarray}{cccc}
         &    1  & 2 & 3\\
\begin{block}{c(ccc)}
  1 & x^1_1      & x^2_1 & x^3_1  \\
  2 & x^1_2      & x^2_2 & x^3_2 \\
  3 & x^1_3      & x^2_3 & x^3_3\\
  4 & x^1_4      & x^2_4 & x^3_4 \\
  5 & \text{NaN} & x^2_5 & x^3_5 \\
  6 & \text{NaN} & x^2_6 & x^3_6 \\
  7 & x^1_7      & x^2_7 & x^3_7 \\
  8 & x^1_8      & x^2_8 & x^3_8 \\
  9 & x^1_9      & x^2_9 & x^3_9 \\
\end{block}
\end{blockarray}
\hspace{0.2cm},\hspace{1cm}
Y=
\begin{blockarray}{cc}
\begin{block}{c(c)}
  1 & x^1_1  \\
  2 & x^1_2  \\
  3 & x^1_3 \\
  4 & x^1_4  \\
  5 & \text{NaN}  \\
  6 & \text{NaN}  \\
  7 & x^1_7  \\
  8 & x^1_8  \\
  9 & x^1_9  \\
\end{block}
\end{blockarray}\ .
 \]

Here, the values for $x^1_5$ and $x^1_6$ are missing and the goal is to estimate their values using an optimized MLP. Then, the lagged table for, e.g.,  lag $l=1$   becomes 

\[\text{Input}=
\begin{blockarray}{ccccccc}
         &    1  & 2 & 3   &    4  & 5 & 6\\
\begin{block}{c(cccccc)}
  1 & x^1_1      & x^2_1 & x^3_1 & x^1_2      & x^2_2 & x^3_2 \\
  2 & x^1_2      & x^2_2 & x^3_2 & x^1_3      & x^2_3 & x^3_3 \\
  3 & x^1_3      & x^2_3 & x^3_3 & x^1_4      & x^2_4 & x^3_4\\
  4 & x^1_4      & x^2_4 & x^3_4 & \text{NaN} & x^2_5 & x^3_5 \\
  5 & \text{NaN} & x^2_5 & x^3_5 & \text{NaN} & x^2_6 & x^3_6\\
  6 & \text{NaN} & x^2_6 & x^3_6 & x^1_7      & x^2_7 & x^3_7 \\
  7 & x^1_7      & x^2_7 & x^3_7 & x^1_8      & x^2_8 & x^3_8 \\
\end{block}
\end{blockarray}
\hspace{0.2cm},\hspace{1cm}
\text{Output}=
\begin{blockarray}{cc}
\begin{block}{c(c)}
  1 & x^1_3 \\
  2 & x^1_4  \\
  3 & \text{NaN}  \\
  4 & \text{NaN}  \\
  5 & x^1_7  \\
  6 & x^1_8  \\
  7 & x^1_9  \\
\end{block}
\end{blockarray}.
 \]

After the lagged tables  have been constructed, we determine if they contain any missing values (NaN). Any  row that  contains at least one NaN entry in the input or output can be removed because the rows in the lagged table are independent. Thus, for the above example, we obtain the following reduced inputs and outputs
{\small 
\[\text{Reduced Input}=
\begin{blockarray}{ccccccc}
         &    1  & 2 & 3   &    4  & 5 & 6\\
\begin{block}{c(cccccc)}
  1 & x^1_1 & x^2_1 & x^3_1 & x^1_2 & x^2_2 & x^3_2 \\
  2 & x^1_2 & x^2_2 & x^3_2 & x^1_3 & x^2_3 & x^3_3 \\
  7 & x^1_7 & x^2_7 & x^3_7 & x^1_8 & x^2_8 & x^3_8 \\
\end{block}
\end{blockarray}
,
\text{Reduced Output}=
\begin{blockarray}{cc}
\begin{block}{c(c)}
  1 & x^1_3 \\
  2 & x^1_4  \\
  7 & x^1_9  \\
\end{block}
\end{blockarray}\ .
 \]
 }
Once the  reduced lagged input and output are obtained, they  are divided into  training and validation datasets. For any architecture,  the reduced lagged training dataset is used  to train the corresponding  neural network model. Since this training is done by stochastic gradient descent (and thus the final trained model can be interpreted as the realization of a stochastic process), we train the model  $k$ times, each with a  different random number  seed. Each of the $k$ trained models is validated against  the  reduced lagged validation dataset, and the architecture's performance  is calculated as the  sample average over all $k$ values. 

Finally, we  impute the missing values sequentially starting from the first missing value to the last missing value, using the previous estimated missing value  to impute the next missing value. In the small example above, we impute the first missing value $x^1_5$ and denote it as $\hat{x}^1_5$ by using $(x^1_3,  x^2_3, x^3_3, x^1_4, x^2_4, x^3_4)$. Then, the next missing value ($x_6^1$) is  estimated based on  $\hat{x}^1_5$, i.e.,   $(x^1_4, x^2_4, x^3_4, \hat{x}^1_5, x^2_5, x^3_5)$. In this way, all missing values are estimated.

\section{Numerical Experiments}
\label{sec: exp}
In this section, we demonstrate that our  proposed method for gap filling of multivariate time series data is general enough to be  used in different applications that have varying observation frequencies and data characteristics. 
We  describe three time series datasets for which we assume that the target variable has one large gap that must be filled. These time series include observed groundwater levels, simulated soil moisture, and derived net ecosystem exchange from flux tower measurements. Below, we  provide details of the  setup of the numerical experiments and alternative gap filling algorithms that we use for comparison.

\subsection{Description of the Datasets }

\subsubsection{Measured groundwater levels}

\begin{figure}[htbp]
    \centering
    \includegraphics[width=1.0\textwidth]{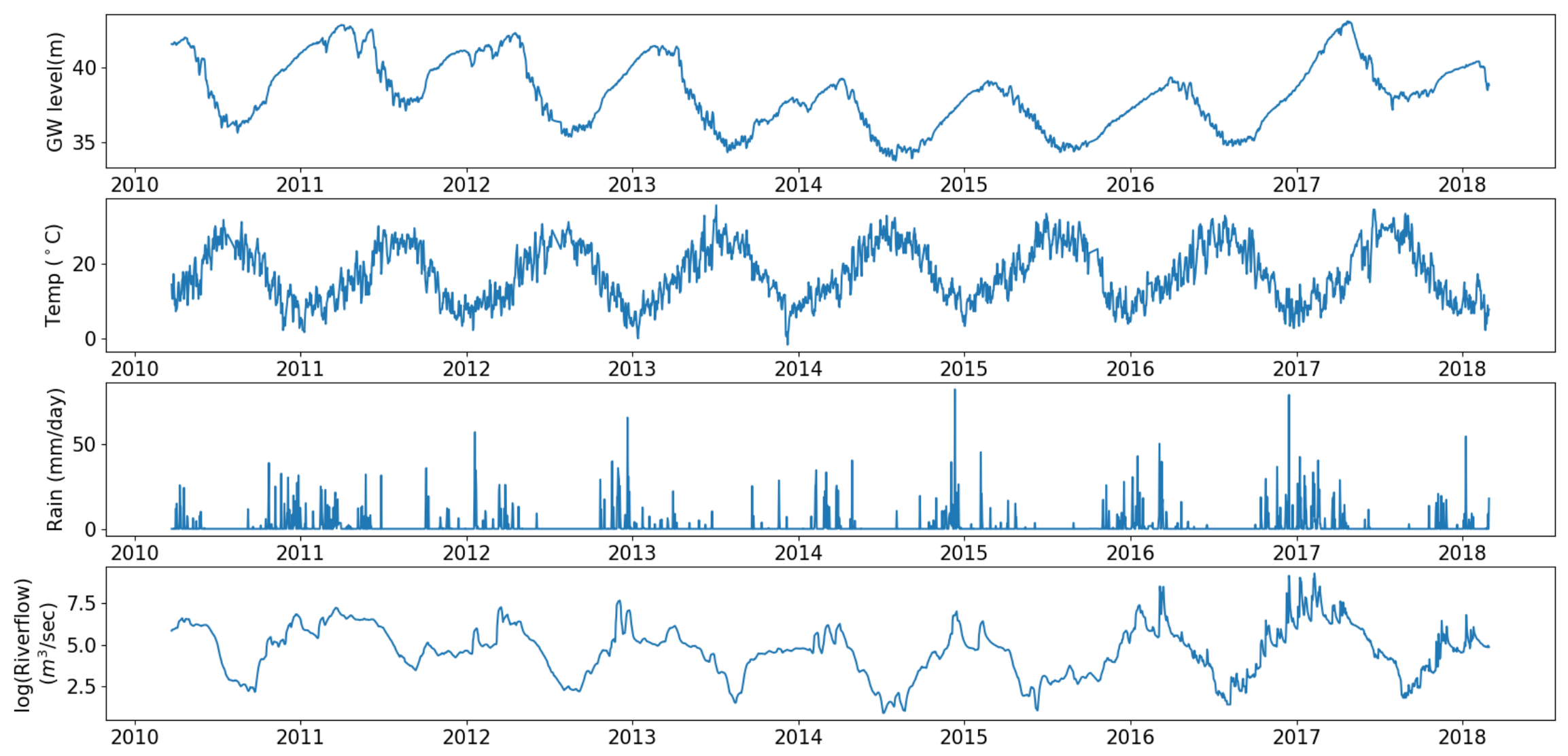}
    \caption{Time series plots for the groundwater level imputation test case. Shown are (top to bottom) the daily groundwater levels (GW), temperature (Temp), precipitation (Rain), and log-transformed riverflux. 
 }
    \label{fig:gw}
\end{figure}

Our first dataset contained groundwater level time series measured in  a groundwater monitoring well  in Butte County in California (CA), United States. 
The monitoring well \footnote{22N01E28J001M, California Natural Resources Agency, Periodic Groundwater Level Measurements, \url{https://data.cnra.ca.gov/dataset/periodic-groundwater-level-measurements}}  is 200 m deep from the surface, and provides the groundwater elevation above the mean sea level at daily time resolution. We obtained the data  from the California Natural Resources Agency.  Our goal was to gap fill the time series of groundwater levels using measurements of the supporting variables -  temperature, precipitation and river discharge. The nearest discharge monitoring station (Butte Creek Durham) measures the daily discharge rate at the Butte Creek about 8 km from the well\footnote{California Data Exchange Center, Butte Creek Durham station, \url{https://cdec.water.ca.gov/webgis/?appid=cdecstation\&sta=BCD}}. Temperature and precipitation data were obtained from the Chico weather station located 7 km from the well site\footnote{California Data Exchange Center, Chico Station, \url{https://cdec.water.ca.gov/webgis/?appid=cdecstation\&sta=CHI}}. The dataset used in the study consisted of approximately eight years (2010-2018) of daily observations. For a  more detailed  description of the dataset and the data sources, we refer the reader to \cite{muller2019surrogate,sahu2020impact}. Figure \ref{fig:gw} shows the time series plots for all variables.

\subsubsection{Simulated soil moisture }

Our second dataset was created with the HYDRUS-1D  model \cite{hydrusmanual, vsimuunek2008modeling}, a soil hydrologic simulation model. The HYDRUS-1D model is a modular, freely-available, state-of-the-art, and widely-used soil hydrologic model with many advanced and coupled water and reactive chemical transport features including equivalent continuum and dual permeability modeling approaches for preferential flow and transport. In particular, the model simulates water, heat and solute transport in variably-saturated and saturated porous media.  The model has been extensively applied for modeling across scales -- from laboratory cores to watersheds \cite{arora2015integrated, baek2020assessment}.

\begin{figure}[htbp]
    \centering
    \includegraphics[width=1.0\textwidth]{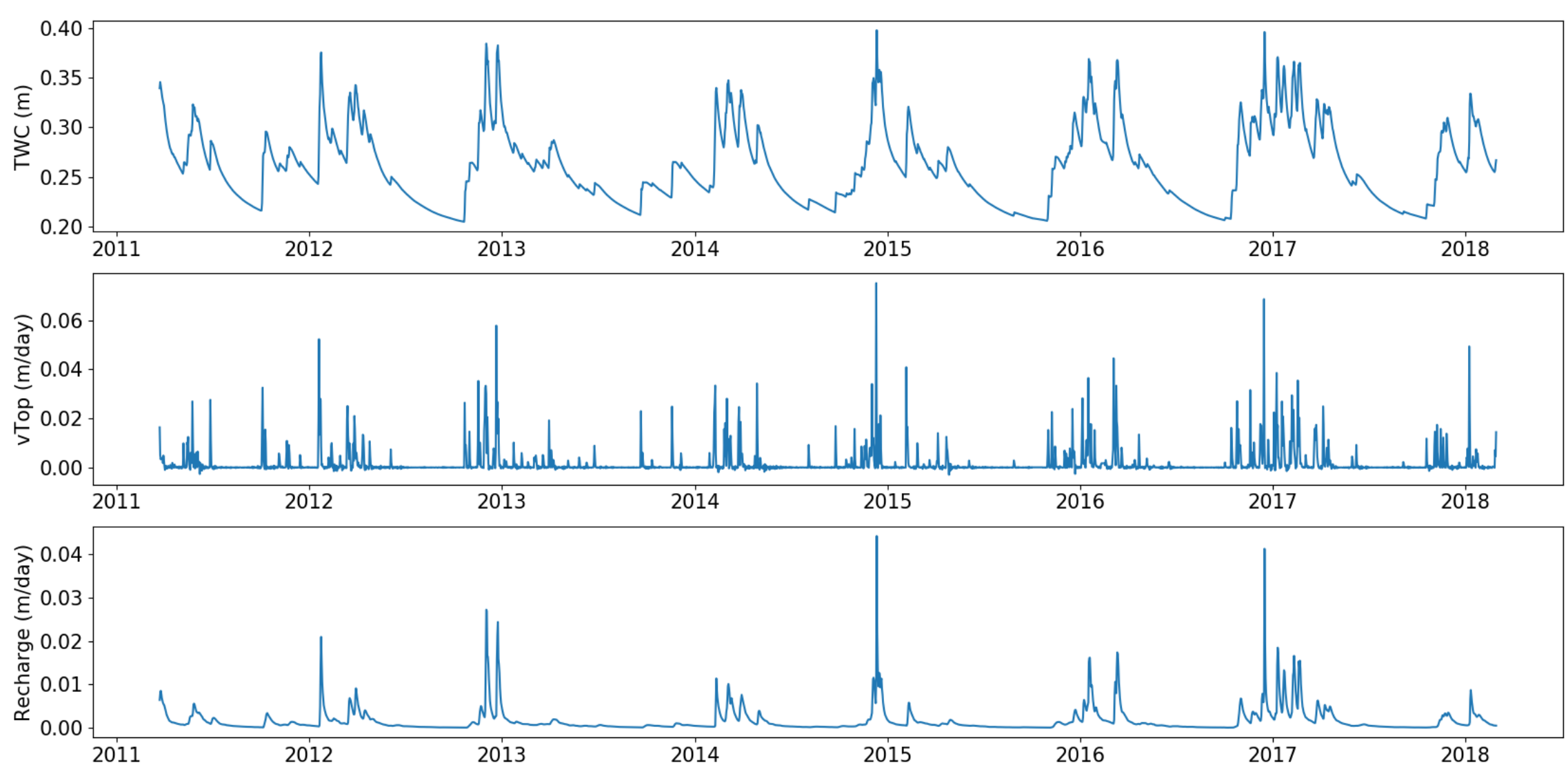}
    \caption{Time series plots for the soil moisture imputation test case. Shown are (top to bottom) the daily total water content (TWC),  the net amount of water entering the soil column from the top (vTop), and the groundwater recharge.
 }
    \label{fig:hydrus}
\end{figure}

  Here, we used the HYDRUS-1D model to simulate a representative soil column in the Butte County in  California to determine the changes in soil moisture profiles over approximately seven years (2011-2018). As a result, our target variable was the daily total water content (TWC) in the upper 80~cm of the simulated soil column. The net amount of water entering the soil column from the top (vTop) and the groundwater recharge were used as the supporting variables for calculations. Figure \ref{fig:hydrus} shows the time series plots for all variables. 
  Because simulation models usually do not lead to missing values, we use it here  as an ideal case study to investigate the general applicability of our MLP-based gap filling approach. 

\subsubsection{Net ecosystem exchange data}
Our third example application was  the most recently produced FLUXNET dataset, that is FLUXNET2015 dataset. It  includes data on CO$_2$, water, energy exchange, and other meteorological and biological measurements \cite{pastorello2020fluxnet2015, FLUXNET2015}. The eddy-covariance method is used to  allow for the non-destructive estimation of fluxes between atmosphere and biosphere for a single site to global scale. FLUXNET datasets have been used in a wide range of research areas ranging from soil microbiology to validation of large-scale earth system models.

\begin{figure}[htbp]
    \centering
    \includegraphics[width=1.0\textwidth]{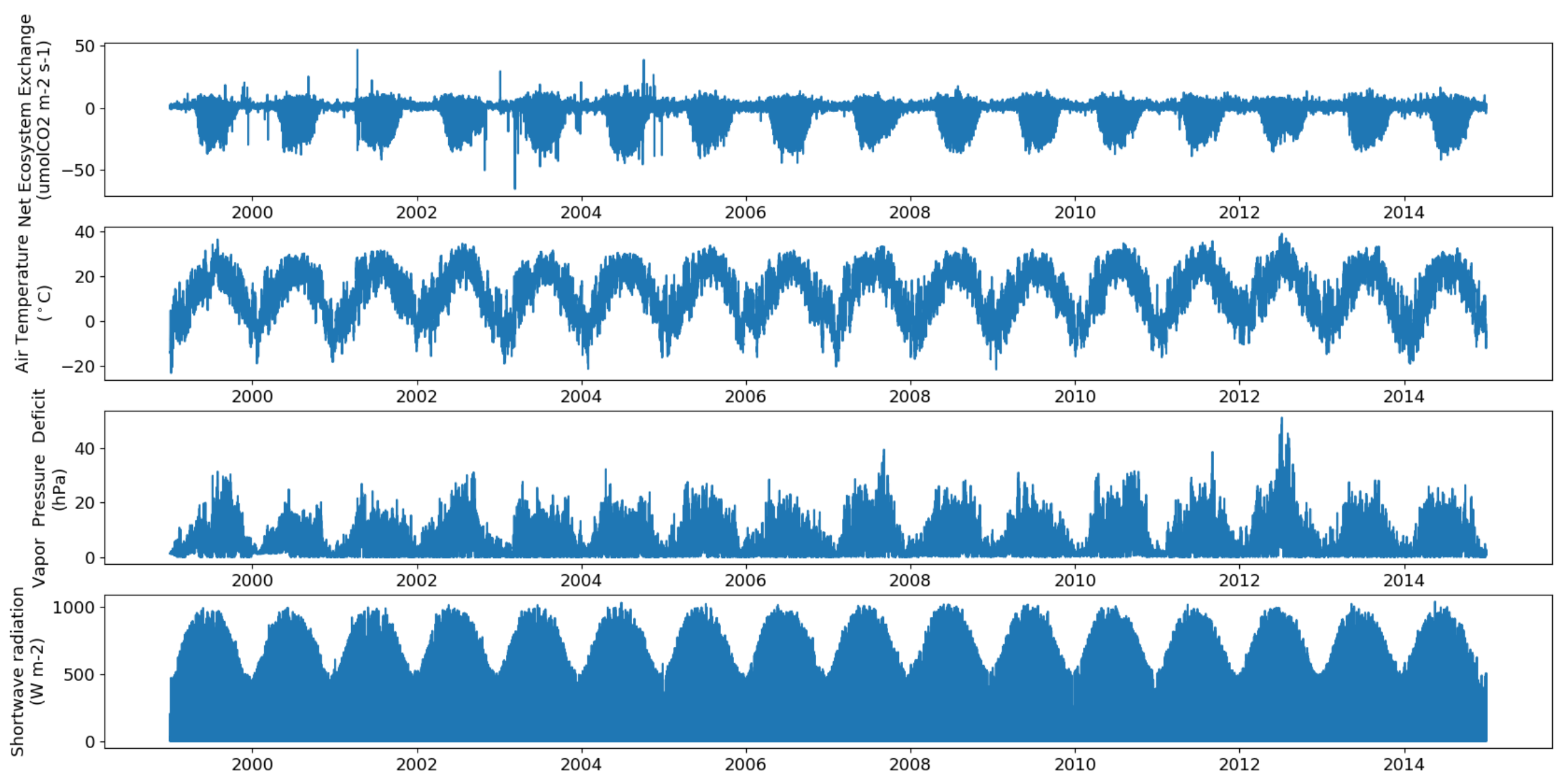}
    \caption{Time series plots for the net ecosystem exchange test case. Shown are (top to bottom) the hourly Net Ecosystem Exchange (NEE\_VUT\_USTAR50), Air Temperature (TA\_F),  Vapor Pressure Deficit (VPD\_F), and incoming shortwave radiation (SW\_IN\_F).
 }
    \label{fig:flux}
\end{figure}

The target and supporting variables for  the Morgan Monroe State Forest site, Indiana, United States for 16 years (1999-2014) \cite{novick2016fluxnet2015} are described in Table \ref{tab:flux_var} and the time series are illustrated in Figure~\ref{fig:flux}. Note that none of the variables in the studied date range  have missing values.\\

\begin{table}[htbp]
  \centering
  \caption{Target and supporting variables with description of FLUXNET2015 dataset}
      \resizebox{\textwidth}{!}{  \begin{tabular}{rccp{14.75em}}
    \toprule
          & \multicolumn{1}{c}{Variable} & \multicolumn{1}{c}{Units} & \multicolumn{1}{c}{Description} \\
    \midrule
    \multicolumn{1}{l}{Target variable} & NEE\_VUT\_USTAR50 & $\mu$molCO2/m$^2$s & Net Ecosystem Exchange, using Variable Ustar Threshold (VUT) for each year, from 50 percentile of USTAR threshold \\
    \midrule
    \multicolumn{1}{l}{Supporting variables} & TA\_F & $^\circ$C& Air temperature \\ \cmidrule{2-4}
          & VPD\_F & hecto Pascals   & Vapor Pressure Deficit \\ \cmidrule{2-4}
          & SW\_IN\_F & Watt/m$^2$ & Shortwave radiation, incoming\\
          \bottomrule
    \end{tabular}%
    }
  \label{tab:flux_var}%
\end{table}%



 In all three  case studies, the supporting variables  had been identified with the help of domain science experts, but filter methods, recursive feature elimination, or sensitivity analyses~\cite{sahu2020impact}    may prove useful to further downselect the most important features.
For all  test cases, temporal information such as the time of day, the day, and the month of the year the  measurements were taken were also included as additional supporting variables. Also, all values of each feature were rescaled with the min-max normalization, i.e., for each  variable $x^i$,  to transform all variables to have the same scale. We found the  minimum value $x^i_{\text{min}}$ and the maximum value $x^i_{\text{max}}$, and we rescaled the  values according to  $\tilde{x}^i = \frac{x^i -x^i_{\text{min}}}{x^i_{\text{max}}-x^i_{\text{min}}}$.

\subsection{Setup of Numerical Experiments}

\subsubsection{HPO details}
We implemented the HPO algorithm for gap filling described in Section~\ref{sec:HPO} in python ($version~3.7$) using PyTorch ($version~1.4.0$) \cite{paszke2017automatic}. All experiments were run on Ubuntu 16.04 with Intel\textsuperscript{\tiny\textregistered} Xeon(R) CPU E3-1245 v6 @ 3.70GHz $\times$ 8, and 31.2 GB memory. We used the same parameter settings as in \cite{muller2019surrogate} and 
we optimized over the following hyperparameters and their ranges:

\begin{itemize}[noitemsep]
    \item batch size $\in\{10,15,\ldots,195, 200\}$,
    \item epochs $\in\{50,100,\ldots,450,500\}$,
    \item layers $\in\{1,2,\ldots,5,6\}$,
    \item number of hidden nodes $\in\{5,10,\ldots,45,50\}$,
    \item dropout rate $\in\{0.0,0.1,\ldots,0.4,0.5\}$,
    \item lags $\in\{30,35,\ldots,360,365\}$. 
\end{itemize}
The optimization hyperparameters and their possible values were the same for  all three test cases.
Note that all parameters were mapped to consecutive integers for the optimization, and  there is a finite number of possible combinations (architectures). With the provided options, there are 7,588,800 possible MLP architectures, and it is computationally intractable to train that many networks, thus motivating the use of automated HPO.

We assumed that each hidden network layer has the same  number of hidden nodes and the same  dropout rate. We use the MSE as performance measure in  tuning the hyperparematers and we split the data into  85\%   training data  and 15\%  validation data.

In the HPO algorithm, we set $n_0=10$ as the initial experimental design size (we used 10 different hyperparameter sets and we trained the models for each to obtain the  model performance, which we use to initialize the surrogate model). We stopped the algorithm after $n = 50$ hyperparameter sets had been tried. In order to take into account the stochasticity that arises from using the stochastic gradient descent method  for training the models,  we computed the hyperparameter performance  as the sample average over multiple training  trials. For example, for the daily groundwater level  application and the HYDRUS-1D application, we trained the model for each  hyperparameter  set  five times using  the same training dataset,  and the performance was defined as the average MSE over those trials. Due to the computational expense associated with training the MLP models for the FLUXNET data (because of the large amount of training data), we computed the performance  only over one training trial for each hyperparameter set.

\subsubsection{Description of gaps in  time series data}
In order to test our developed gap filling method, we  created an artificial continuous gap in the target time series. We studied different instances in which the gaps have varying lengths and are located at different times of the year. By using artificial gaps,  we have the ability to compare the gap-filled values to the true held-out observations as well as analyze and compare the performance of different gap filling methods. In all three test cases, the supporting variables did not have gaps.

For the groundwater application, we created gaps of three, six and twelve  months of daily observations. For each case, we used  12 different problem instances in which we removed the data of different months of the year, which allowed us to analyze a possible dependence of the performance of the gap filling methods on  the trend characteristics of the missing data.  For example, for the three month gap, we assumed that  groundwater levels were missing   (i) from January through March 2012,  (ii) from February through April 2012, (iii) from March through May 2012, etc.
For the  HYDRUS-1D application, we assumed that the total water content values were missing for 12 and 15 months, respectively. 
For the 12-month gap, we assumed that the total water content values were missing   (i) from January 2013 through December 2013,  (ii) from February 2013 through January 2014, and (iii) from March 2013 through February 2014. For the 15 months gap, we considered 9 cases where  the missing values were staring from the months of April, May, till December.
 Finally, for the hourly FLUXNET data, we considered between 2,159 hours (3 months) and 35,063 hours (36 months) of missing values in the Net  Ecosystem  Exchange data (NEE\_VUT\_USTAR50).

\subsubsection{Gap filling algorithms used in the numerical  comparison}

We compared our proposed gap filling method with state-of-the-art imputation packages, including  \textsf{ImputeTS} \cite{moritz2017imputets} and \textsf{mtsdi} \cite{junger2018package}.  \textsf{ImputeTS} is a univariate time series missing value imputation package implemented in R. We used the \texttt{na.seadec} function with an interpolation algorithm. 
The \texttt{na.seadec} function imputes missing values with deseasonalized time series data and adds the seasonality again.
It has the option to identify the number of observations before the seasonal pattern repeats. However, in our applications, this feature did not yield satisfactory results, and thus we manually set the daily time series frequency, i.e., the frequency of 365. 
\textsf{mtsdi} is an expectation-maximization algorithm based multivariate time series missing value imputation package in R. We used the default settings, i.e.,  the \texttt{mnimput} function with the spline method, and we set the spline smooth control to 7. With \textsf{mtsdi}, we included indicator variables such as the  month, week, and day.\\

\subsection{Results and Discussion}

\subsubsection{Groundwater level predictions}
Table \ref{tab:groundwater_3month}
presents a  comparison of the root mean squared error (RMSE)  computed between the predicted values and the true (held-out) values for the groundwater use cases with  three, six, and  twelve months of missing daily data values, respectively. Shown are the RMSEs of our proposed MLP-based gap filling method, \texttt{ImputeTS},  and \texttt{mtsdi}.  The RMSE is given in meters (m). We show the optimal hyperparameters obtained with our method when using  the Gaussian process  (GP) and the radial basis function (RBF) as surrogate model, respectively.

\begin{figure}[htbp]
    \centering
    \includegraphics[width=1.0\textwidth]{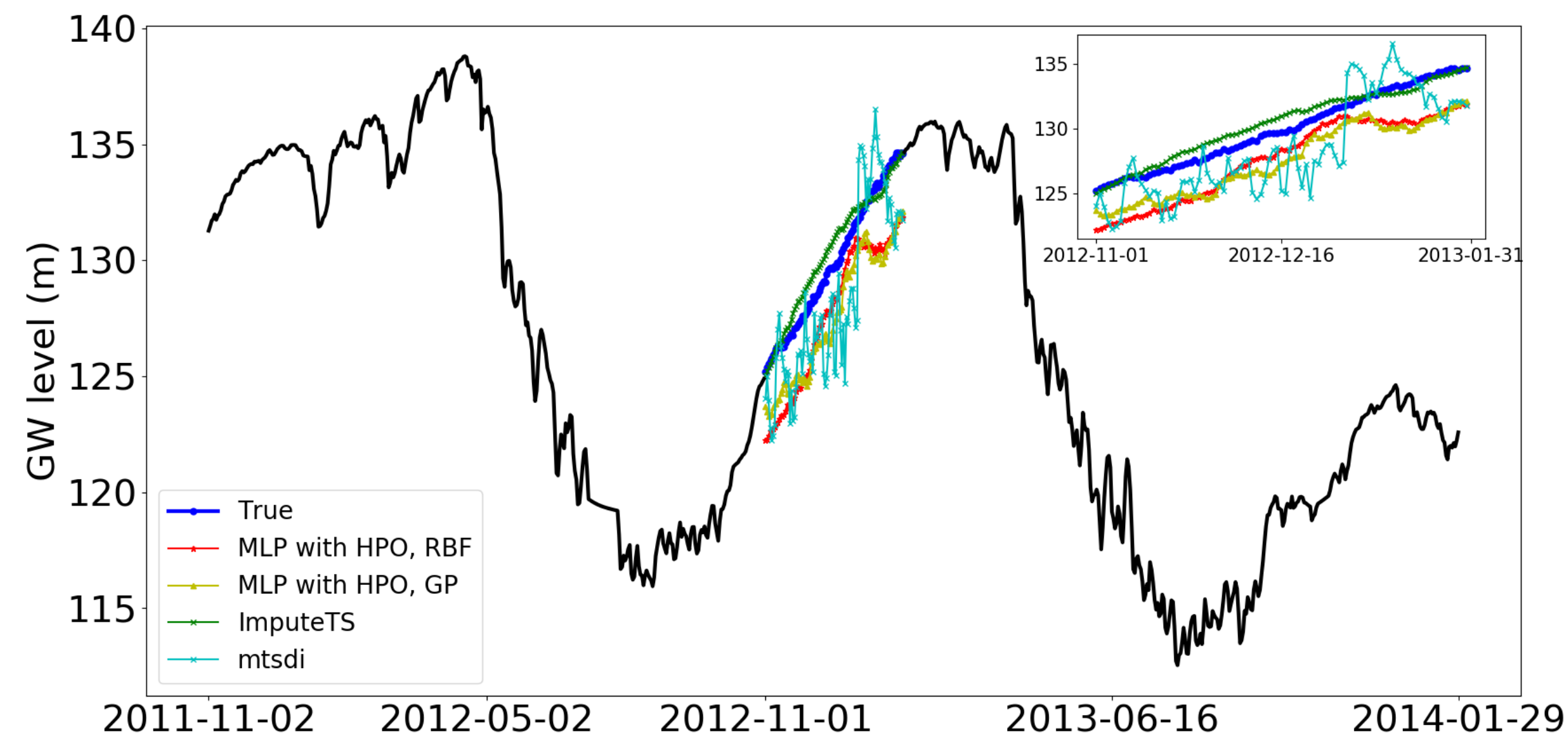}
    \caption{Daily groundwater level imputation for \textit{three} months of missing daily measurement data (November 2012 - January 2013). The missing data have a linear trend which is captured by \textsf{imputeTS} and the MLP-based method.}
    \label{fig:gw_3month}
\end{figure}

For three months of missing daily observations  (top section of Table \ref{tab:groundwater_3month}), our results show that the lowest RMSE was achieved by different methods depending on the date range for which the data are missing.  
For example, for the three months that span November 2012-January 2013 (see also Figure \ref{fig:gw_3month}), the groundwater levels show a linear growth.   All gap filling methods  captured this increasing trend and \texttt{ImputeTS}  yielded the best performance in terms of RMSE. This is due to the fact that  \texttt{ImputeTS} uses \texttt{na.seadec} with an interpolation algorithm. In other words, \texttt{ImputeTS} performed seasonally decomposed missing value imputation  by linear interpolation which yielded excellent agreement with the held-out data for the missing data segment that had linear behavior.
Also the MLP-based gap filling methods were able to capture the linear trend, but the predictions were not as accurate. \textsf{mtsdi} introduced strong oscillations around the overall growing trend. 
Moreover, \textsf{mtsdi} performed overall the worst for all date ranges for  the three month test case.
\begin{table}[htbp]
  \centering

  \caption{RMSE values (m) for the different gap filling methods for \textit{three, six}, and \textit{twelve} months of missing values in daily groundwater levels and optimal hyperparameters ([batch size, \# epochs, \# layers, \# nodes per layer, dropout rate, \# lags])  for both RBF- and GP-based HPO.\protect\footnotemark}
   \resizebox{\textwidth}{!}{\begin{tabular}{ccc|c|c|c|ccc|c}
\cmidrule{1-7}\cmidrule{9-10}    \multicolumn{2}{c}{Missing date range (yyyy-mm-dd)} & \#obs & \multicolumn{4}{c}{RMSE}      &       & \multicolumn{2}{c}{Hyperparameters} \\
\cmidrule{4-7}\cmidrule{9-10}    From & To &       & MLP RBF & MLP GP & ImputeTS & mtsdi &       & MLP RBF & MLP GP \\
\cmidrule{1-7}\cmidrule{9-10}    
    2012-01-01 & 2012-03-31 & 91 & \textbf{0.34}  & 0.38  & 0.56  & 1.23  &       & [145,  150,    1,   45,    0,  330]  &  [190,  450,    4,   50,    0,  270] \\
    2012-02-01 & 2012-04-30 & 90 & \textbf{0.45}  & 0.56  & 0.49  & 1.18  &       & [120,  400,    1,   25,    0,  250]  &  [95,  450,    3,   40,    0,  210] \\
    2012-03-01 & 2012-05-31 & 92 & \textbf{0.54}  & 0.61  & 0.97  & 1.34  &       & [195,  400,    1,   45,    0,  250] &  [170,  350,    2,   20,    0,  290] \\
    2012-04-01 & 2012-06-30 & 91 & \textbf{0.56}  & 0.62  & 0.72  & 1.38  &       & [175,  450,    1,   40,    0,  240]  &  [70,  500,    1,   25,    0,  345] \\
    2012-05-01 & 2012-07-31 & 92 & \textbf{0.61}  & 0.62  & \textbf{0.61}  & 1.17  &       & [130,  450,    4,   45,    0,  330]  &  [170,  350,    1,   30,    0,  240] \\
    2012-06-01 & 2012-08-31 & 92 & 0.42  & 0.41  & \textbf{0.39}  & 1.64  &       & [170,  250,    1,   45,    0,  265]  &  [70,  500,    1,   25,    0,  345] \\
    2012-07-01 & 2012-09-30 & 92 & \textbf{0.26}  & 0.33  & 0.35  & 1.74  &       & [170,  250,    1,   45,    0,  265]  &  [70,  500,    1,   25,    0,  345] \\
    2012-08-01 & 2012-10-31 & 92 & \textbf{0.35}  & \textbf{0.35}  & 0.49  & 1.56  &       & [170,  250,    1,   45,    0,  265]  &  [70,  500,    1,   25,    0,  345] \\
    2012-09-01 & 2012-11-30 & 91 & 0.49  & 0.80  & \textbf{0.28}  & 1.02  &       & [125,  400,    2,   45,    0,  230]  &  [50,  150,    4,   50,    0,  290] \\
    2012-10-01 & 2012-12-31 & 92 & 0.80  & 0.93  & \textbf{0.25}  & 0.75  &       & [170,  250,    1,   45,    0,  265]  &  [70,  500,    1,   25,    0,  345] \\
    2012-11-01 & 2013-01-31 & 92 & 0.74  & 0.75  & \textbf{0.24}  & 0.73  &       & [170,   250,     1,    45,     0,   265]  &  [70,   500,     1,    25,     0,   345] \\
    2012-12-01 & 2013-02-28 & 90 & 0.69  & 0.67  & \textbf{0.38} & 0.99  &       & [135,   450,     3,    45,     0,   305]  &  [110,   200,     1,    10,     0,   255] \\ \hline
    2012-01-01 & 2012-06-30 & 181   & 0.54  & \textbf{0.48}          &       0.62 & 4.80   &              &   [145. 400.   5.  45.   0. 265.]  &   [185. 450.   1.  40.   0. 250.] \\
    2012-02-01 & 2012-07-31 & 181   & \textbf{0.44}  & 0.45  &       0.60 & 4.10       &       &   [140. 400.   2.  45.   0. 310.]  &   [ 50. 200.   3.  40.   0. 310.] \\
    2012-03-01 & 2012-08-31 & 183   & \textbf{0.46}  & 0.47  &       0.82 & 5.63       &       &   [190. 500.   1.  50.   0. 325.]  &   [190. 450.   4.  50.   0. 270.] \\
    2012-04-01 & 2012-09-30 & 182   & \textbf{0.48}  & 0.91  &       0.55 & 7.35       &       &   [155. 400.   2.  45.   0. 245.]  &   [200. 400.   4.  45.   0. 285.] \\
    2012-05-01 & 2012-10-31 & 183   & \textbf{0.57}  & 0.60  &       1.05 & 8.69       &       &   [195. 450.   1.  40.   0. 335.]  &   [ 50. 250.   1.  45.   0. 320.] \\
    2012-06-01 & 2012-11-30 & 182   & \textbf{0.54}  & 0.70  &       0.76 & 8.84       &       &   [ 90. 400.   1.  45.   0. 310.]  &   [145. 500.   4.  20.   0. 245.] \\
    2012-07-01 & 2012-12-31 & 183   & 0.80  &0.77  &       \textbf{0.70} & 6.67       &       &   [ 85. 250.   1.  45.   0. 305.]  &   [ 70. 500.   1.  25.   0. 345.] \\
    2012-08-01 & 2013-01-31 & 183   & 0.64  & 0.55  &       \textbf{0.53} & 4.51       &       &   [195. 450.   1.  35.   0. 260.]  &   [160. 400.   1.  35.   0. 245.] \\
    2012-09-01 & 2013-02-28 & 180   & \textbf{0.44}  & 1.12  &       0.45 & 2.92       &       &   [ 65. 450.   4.  45.   0. 215.]  &   [ 60. 400.   4.  25.   0. 285.] \\
    2012-10-01 & 2013-03-31 & 181   & \textbf{0.59}  & 1.03  &       1.17 & 4.67       &       &   [195. 450.   1.  30.   0. 230.]  &   [155. 400.   2.  35.   0. 305.] \\
    2012-11-01 & 2013-04-30 & 180   & 0.85  & \textbf{0.78}  &       1.67 & 8.51       &       &   [120. 300.   1.  25.   0. 275.]  &   [ 60. 250.   4.  50.   0. 255.] \\
    2012-12-01 & 2013-05-31 & 181   & \textbf{0.78} & 0.81  &       1.09 & 8.76       &       &   [145. 400.   5.  45.   0. 265.]  &   [ 80. 150.   2.  35.   0. 255.] \\ \hline
    2012-02-01 & 2013-01-31 & 366 & \textbf{0.71}  & 0.74  & 1.13  & 1.32  &       & [155,   450,     4,    45,     0,   350]  &  [70,   500,     1,    25,     0,   345] \\
    2012-03-01 & 2013-02-28 & 365 & 0.74  &\textbf{ 0.71}  & 0.79  & 1.25  &       & [160,   350,     1,    40,     0,   305]  &  [70,   300,     2,    15,     0,   260] \\
    2012-04-01 & 2013-03-31 & 365 & 0.82  & \textbf{0.73}  & 0.91  & 1.29  &       & [85,   450,     1,    40,     0,   320]  &  [120,   400,     4,    25,     0,   270] \\
    2012-05-01 & 2013-04-30 & 365 & \textbf{0.71}  & 1.10  & 1.27  & 1.35  &       & [185,   450,     1,    45,     0,   255] &  [185,   100,     2,    45,     0,   270] \\
    2012-06-01 & 2013-05-31 & 365 & 0.64  & \textbf{0.62}  & 1.07  & 1.36  &       & [185,   450,     1,    45,     0,   255]  &  [140,   500,     1,    15,     0,   235] \\
    2012-07-01 & 2013-06-30 & 365 & \textbf{0.76}  & 0.78  & 1.08  & 1.31  &       & [185,   450,     1,    45,     0,   255]  &  [85,   400,     5,    25,     0,   240] \\
    2012-08-01 & 2013-07-31 & 365 & \textbf{0.86}  & 0.92  & 1.11  & 1.25  &       & [110,   400,     1,    45,     0,   260]  &  [170,   300,     2,    40,     0,   315] \\
    2012-09-01 & 2013-08-31 & 365 & \textbf{0.85}  & 0.95  & 1.03  & 1.20  &       & [145,   450,     5,    40,     0,   265]  &  [170,   300,     2,    40,     0,   315] \\
    2012-10-01 & 2013-09-30 & 365 & \textbf{0.88}  & \textbf{0.88}  & 1.18  & 1.19  &       & [145,   450,     5,    40,     0,   265]  &  [140,   300,     1,    45,     0,   260] \\
    2012-11-01 & 2013-10-31 & 365 & \textbf{0.82}  & 0.85  & 0.87  & 1.19  &       & [145,   450,     5,    40,     0,   265]  &  [195,   500,     5,    35,     0,   255] \\
    2012-12-01 & 2013-11-30 & 365 & \textbf{0.81}  & 1.02  & 0.85  & 1.17  &       & [185,   450,     1,    45,     0,   255] &  [70,   500,     1,    25,     0,   345] \\
\cmidrule{1-7}\cmidrule{9-10}    
\multicolumn{10}{l}{\small\textbf{Bold} numbers indicate the best value for a given problem instance. The column ``\#obs'' indicates the number of missing observations.}\\ \multicolumn{10}{l}{\small``MLP RBF'' and ``MLP GP'' indicate the methods using the MLP with the RBF and GP model during HPO, respectively.}\\
    \end{tabular}}%

       
  \label{tab:groundwater_3month}%

\end{table}%

\begin{figure}[htbp]
    \centering
    \includegraphics[width=1.0\textwidth]{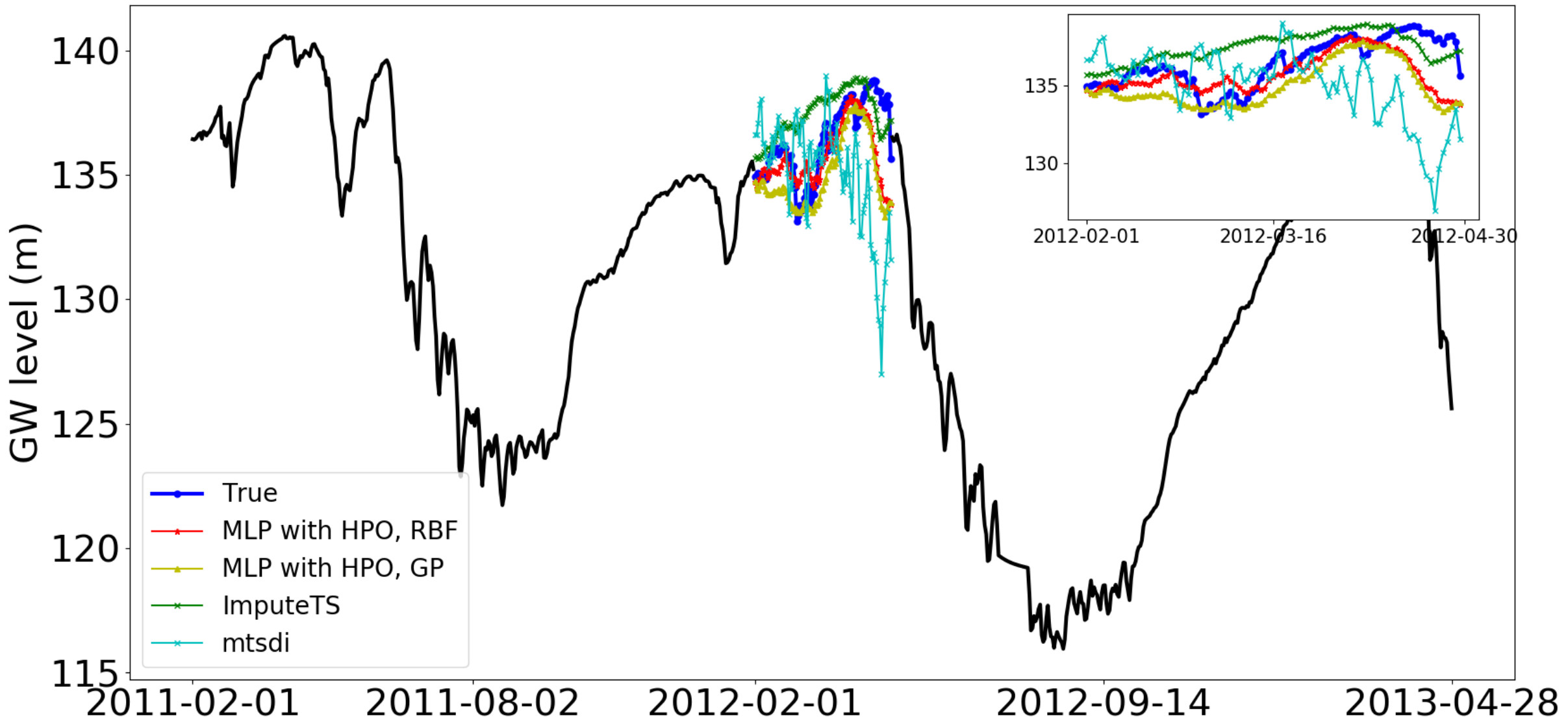}
    \caption{Daily Groundwater level imputation for \textit{three} months of missing daily measurement data (February 2012 - April 2012). The missing data have a nonlinear trend and non-interpolating methods like our MLP approach perform better for these instances.}
    \label{fig:gw_3month1}
\end{figure}

\begin{figure}[htbp]
    \centering
    \includegraphics[width=1.0\textwidth]{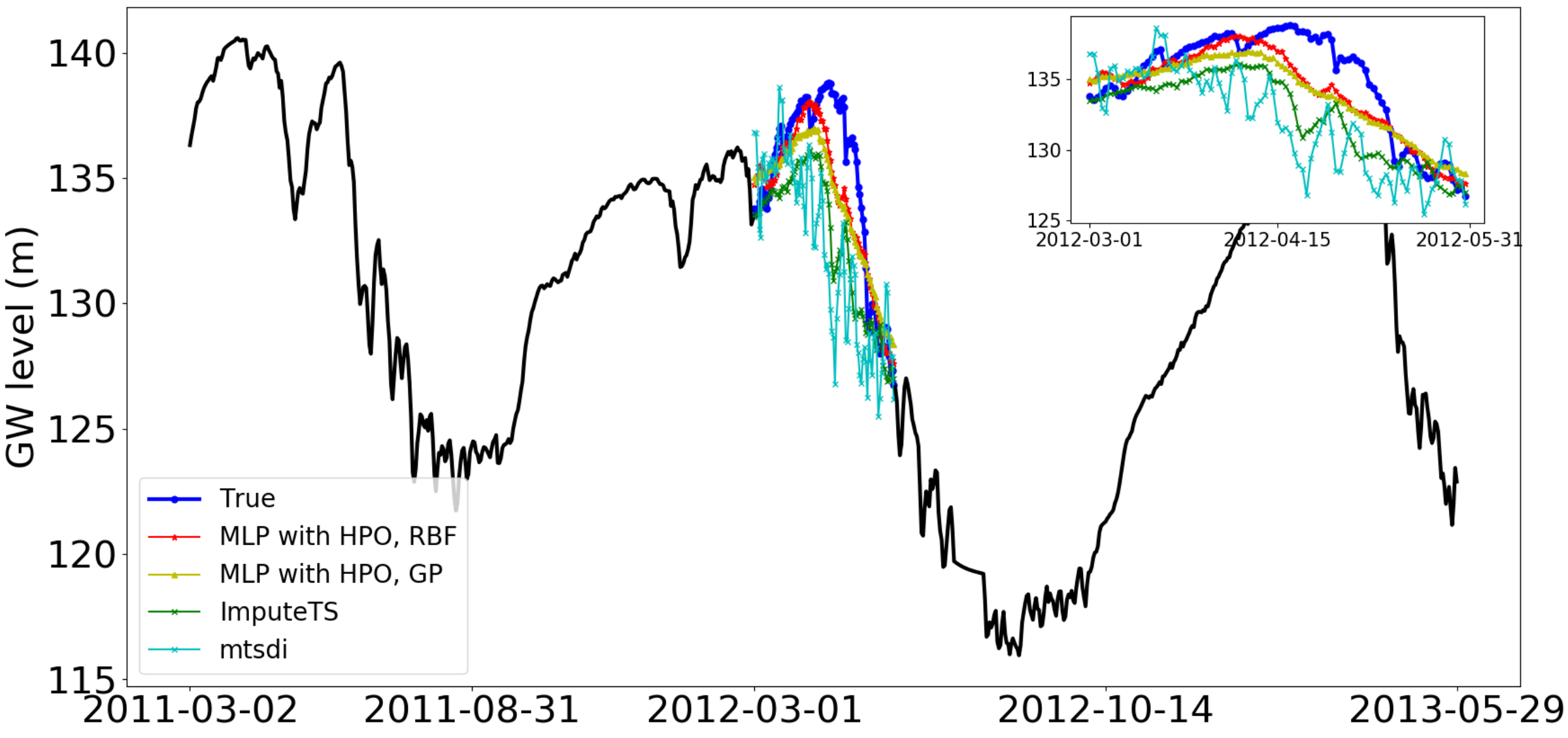}
    \caption{Daily Groundwater level imputation for \textit{three} months of missing daily measurement data (March 2012 - May 2012). The missing data have a nonlinear trend and non-interpolating methods like our MLP approach perform better for these instances.}
    \label{fig:gw_3month2}
\end{figure}

In contrast, Figures~\ref{fig:gw_3month1} and \ref{fig:gw_3month2} show scenarios where the data are  behaving nonlinearly. In Figure~\ref{fig:gw_3month1},  the interpolation used by \textsf{imputeTS} failed to capture the trend and produced an almost flat prediction. Both MLP based methods captured the nonlinear trend in the missing data better, yet the deviation over the last $\sim$ 3 weeks was larger. \textsf{mtsdi} made again predictions with large oscillations around the true data, and for the last 6 weeks of missing data, it deviated significantly from the truth.  

\begin{figure}[htbp]
    \centering
    \includegraphics[width=1.0\textwidth]{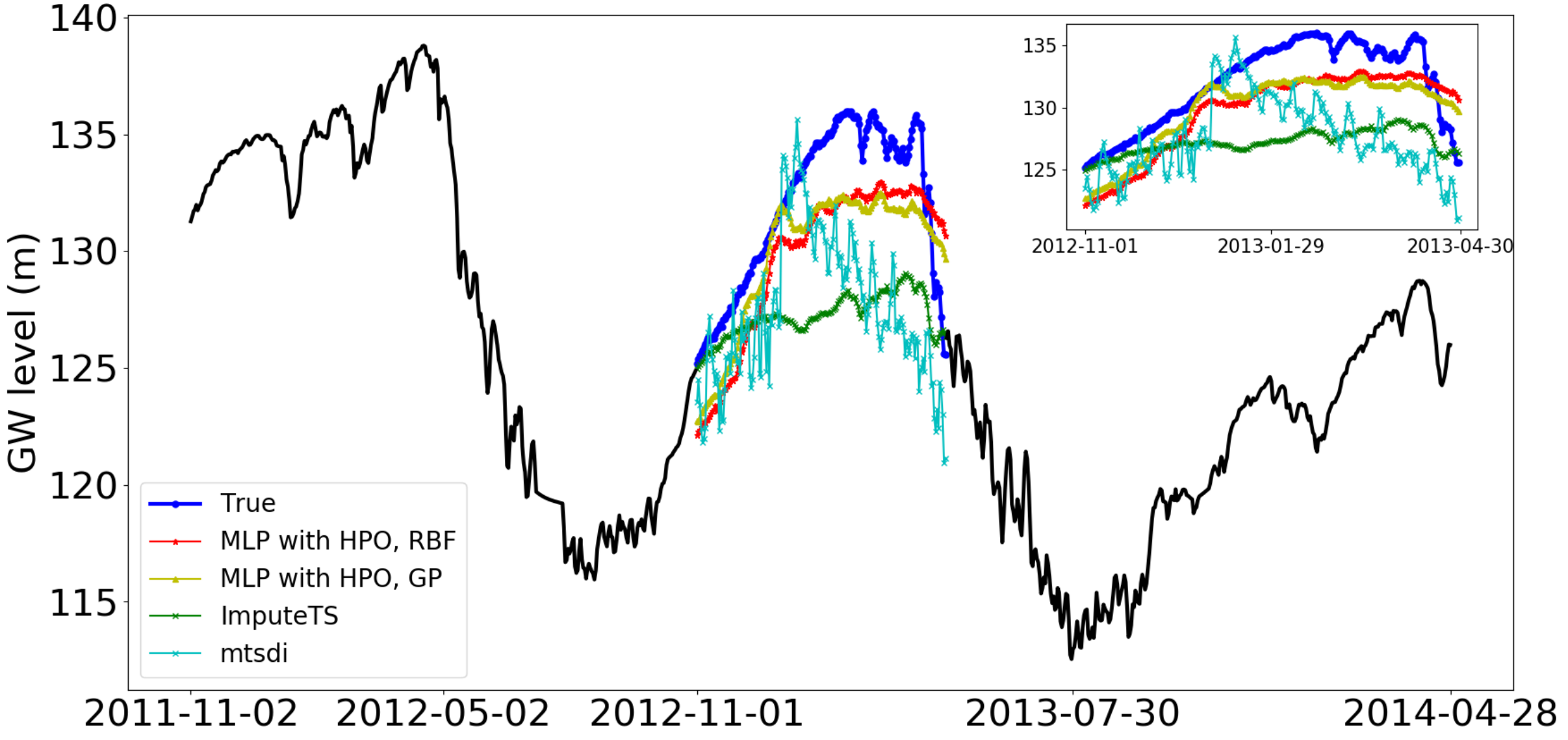}
    \caption{Daily Groundwater level imputation for \textit{six} months of missing daily measurement data (November 2012 - April 2013). }
    \label{fig:gw_6month}
\end{figure}

The results for the case where six months of daily observations were missing are summarized in the middle section of  Table~\ref{tab:groundwater_3month}. Except for two cases (missing values from  July 2012 to December 2012 and from August 2012 to January 2013), our proposed approach with either RBF or GP made more accurate predictions than \textsf{imputeTS} and \textsf{mtsdi} in terms of RMSE. Figure~\ref{fig:gw_6month} is a comparison between all methods with missing values ranging from November 2012 to April 2013. We can see that our proposed method  captured the  high peak better than the other methods and \textsf{imputeTS} made again a fairly flat prediction.
Similarly to the cases of three months of continuous gaps, \textsf{mtsdi} provided the lowest accuracy.  Note that the increased size of the continuous gap allowed for a larger probability that seasonality effects will be  included in the missing value ranges.

\begin{figure}[htbp]
    \centering
    \includegraphics[width=1.0\textwidth]{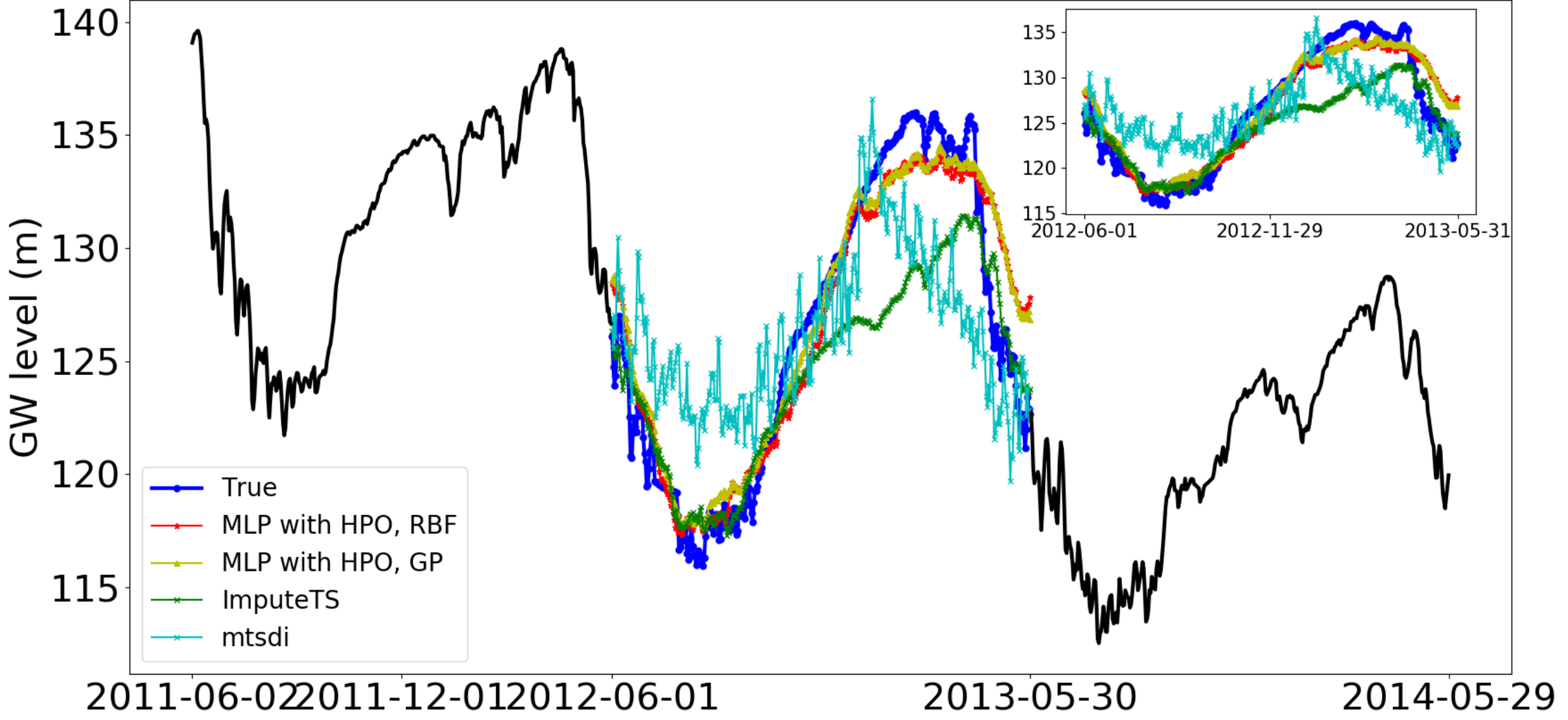}
    \caption{Daily Groundwater level imputation for \textit{twelve} months of missing daily measurement data (June 2012 - May 2013). Although all imputation methods try to capture the nonlinearities, the MLP based imputation agrees best with the true values.}
    \label{fig:gw_12month}
\end{figure}

The results for the  case where 12 months of daily observations were missing are summarized in the lower section of Table~\ref{tab:groundwater_3month} and in Figure~\ref{fig:gw_12month}. The MLP-based gap filling method that uses the RBF as surrogate in HPO  outperformed both \textsf{imputeTS} and \textsf{mtsdi} for all instances. 
 Figure \ref{fig:gw_12month} shows that the MLP-based gap filling method captured the nonlinearities in the missing data significantly better than \textsf{mtsdi} (which introduced strong oscillations and did not capture the extrema of the missing values) and \textsf{imputeTS} (which failed to capture the  winter data  (November 2012-April 2013) and appeared like a piece-wise linear approximation of the missing data). The MLP-based imputation method showed stronger deviations from the true data for the  last $\sim$1.5 months of gap filling, which is possibly due to the fact that our method estimates the missing values sequentially based on the previously estimated values, and thus it is possible that errors started to accumulate. 


Table \ref{tab:groundwater_3month}
also show the optimal hyperparameters for the MLP. The results  indicate that the identified optimal  MLP architectures depended on the date range for which the data were missing, and thus the same hyperparameters may not be optimal for all date ranges and    tuning is necessary.  The RMSE performance of a specific architecture was  impacted by the data that were missing. For example, in Table \ref{tab:groundwater_3month}, we see that the same optimal hyperparameters were obtained for the date ranges June 2012-August 2012, July 2012-September 2012, August 2012-October 2012, October 2012-December 2012, and November 2012-January 2013 when using either the RBF or the GP during hyperparameter tuning, respectively. However, the corresponding RMSEs ranged between 0.26 and 0.8 for RBF, and between 0.33 and 0.93 for the GP. 
We observe a similar behavior for the cases of six and  twelve months of missing data. 

\subsubsection{Total water content predictions}

\begin{figure}[htbp]
    \centering
    \includegraphics[width=1.0\textwidth]{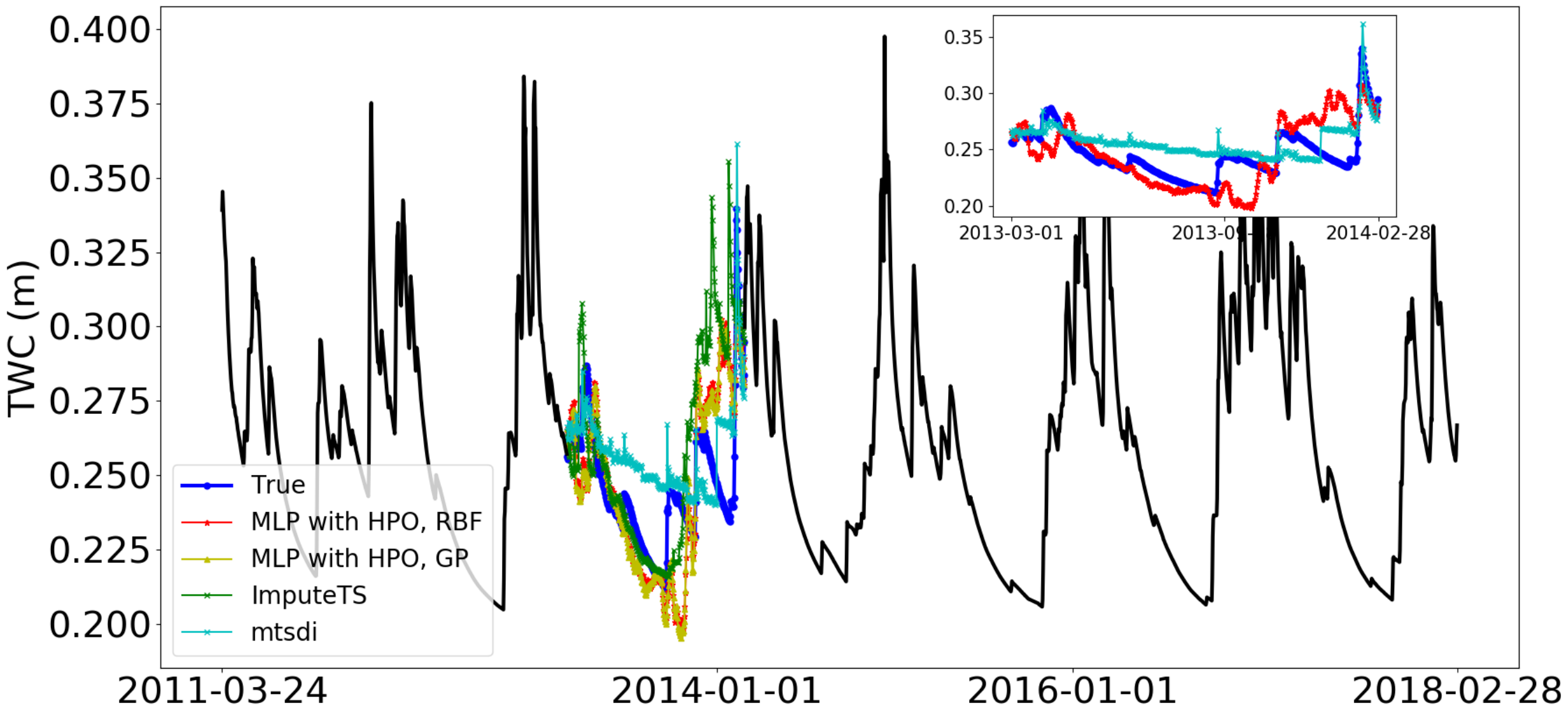}
    \caption{Comparison of gap filling methods for daily TWC with missing values between March 2013 and February 2014.}
    \label{fig:twc_12month}
\end{figure}

Table \ref{tab:twc_12month} shows the results of our data imputation for the  TWC simulated with  HYDRUS-1D. Shown are the RMSE values in meters between the true and estimated values obtained with  the  different  gap  filling  methods for the 12- and 15-month cases. The optimal hyperparameters for the MLP-based approaches are also shown.  Our MLP-based gap filling method achieved lower RMSE values  than \textsf{ImputeTS} for all three tested 12-month date ranges, but it was outperformed by \textsf{mtsdi} for two  test cases. Figure \ref{fig:twc_12month} shows the true simulated TWC values and the values as estimated by the gap filling methods for a 12-month case. We can see that all methods had difficulties capturing  the true data, and  our proposed method and \textsf{imputeTS} did a better job than \textsf{mtsdi} at capturing the first seven months of missing data, but then both methods generated larger errors during the remaining five months.  In contrast, \textsf{mtsdi} did not capture the trend in the data of the first seven months well,  but the resulting RMSE is smaller overall regardless.

\begin{table}[htbp]
  \centering
  \caption{RMSE (in meters) for the different gap filling methods for 12 and 15 months of missing values in daily TWC    and optimal hyperparameters ([batch size, \# epochs, \# layers, \# nodes per layer, dropout rate, \# lags]) for both RBF- and GP-based HPO. }
   \resizebox{\textwidth}{!}{\begin{tabular}{ccc|c|c|c|cccc}
\cmidrule{1-7}\cmidrule{9-10}    \multicolumn{2}{c}{Missing date range (yyyy-mm-dd)} & \#obs & \multicolumn{4}{c}{RMSE}      &       & \multicolumn{2}{c}{Hyperparameters} \\
\cmidrule{4-7}\cmidrule{9-10}    From & To &  & MLP RBF & MLP GP & ImputeTS & mtsdi &       & MLP RBF & GP \\
\cmidrule{1-7}\cmidrule{9-10}    2013-01-01 & 2013-12-31 & 365 & \textbf{0.0084} & 0.0116 & 0.0301 & 0.0155 &       & [50,  500,    1,   50,    0,   45]  &  [55,  300,    6,   50,    0,   55] \\
    2013-02-01 & 2014-01-31 & 365 & 0.0246 & 0.0233 & 0.0425 & \textbf{0.0184} &       & [60,  500,    1,   35,    0,  355]  &  [70,  500,    1,   25,    0,  345] \\
    2013-03-01 & 2014-02-28 & 365 & 0.0225 & 0.0228 & 0.0364 & \textbf{0.0185} &       & [70,  500,    1,   45,    0,  345] &  [ 70,  500,    1,   25,    0,  345] \\ \hline
    2013-04-01 & 2014-06-30 & 456 & 0.0204 & 0.0210 & 0.0287 & \textbf{0.0169} &       & [50,  500,    1,   45,    0,  345] &   [70,  500,    1,   25,    0,  345] \\
    2013-05-01 & 2014-07-31 & 457 & 0.0236 & 0.0202 & 0.0280 & \textbf{0.0181} &       & [195,  450,    1,   40,    0,  335]  &  [70,  500,    1,   25,    0,  345] \\
    2013-06-01 & 2014-08-31 & 457 & 0.0233 & 0.0205 & 0.0276 & \textbf{0.0187} &       & [195,  450,    1,   40,    0,  335] &  [70,  500,    1,   25,    0,  345] \\
    2013-07-01 & 2014-09-30 & 457 & 0.0235 & 0.0209 & 0.0278 & \textbf{0.0197} &       & [195,  450,    1,   40,    0,  335]  &  [70,  500,    1,   25,    0,  345] \\
    2013-08-01 & 2014-10-31 & 457 & 0.0239 & 0.0210 & 0.0273 & \textbf{0.0195} &       & [195,  450,    1,   40,    0,  335]  &  [70,  500,    1,   25,    0,  345] \\
    2013-09-01 & 2014-11-30 & 456 &\textbf{ 0.0053} & 0.0243 & 0.0299 & 0.0178 &       & [50,  500,    1,   50,    0,   30]  &  [ 70,  500,    1,   25,    0,  345] \\
   2013-10-01 & 2014-12-31 &457 & 0.0276 & 0.0282 & 0.0383 & \textbf{0.0235} &       & [195,  450,    1,   40,    0,  335]  &  [70,  500,    1,   25,    0,  345] \\
    2013-11-01 & 2015-01-31 & 457 & 0.0291 & 0.0266 & 0.0472 & \textbf{0.0229} &       & [195,  450,    1,   40,    0,  335]  &  [70,  500,    1,   25,    0,  345] \\
    2013-12-01 & 2015-02-28 & 455 & \textbf{0.0141} & 0.0203 & 0.0308 & 0.0228 &       & [50,  450,    1,   45,    0,  120]  &  [70,  500,    1,   25,    0,  345] \\
    \cmidrule{1-7}\cmidrule{9-10}  
\multicolumn{10}{l}{\small\textbf{Bold} numbers indicate the best value for a given problem instance. The column ``\#obs'' indicates the number of missing observations.}\\ \multicolumn{10}{l}{\small``MLP RBF'' and ``MLP GP'' indicate the methods using the MLP with the RBF and GP model during HPO, respectively.}\\    
    \end{tabular}}%
  \label{tab:twc_12month}%
\end{table}%

\begin{figure}[htbp]
    \centering
    \includegraphics[width=1.0\textwidth]{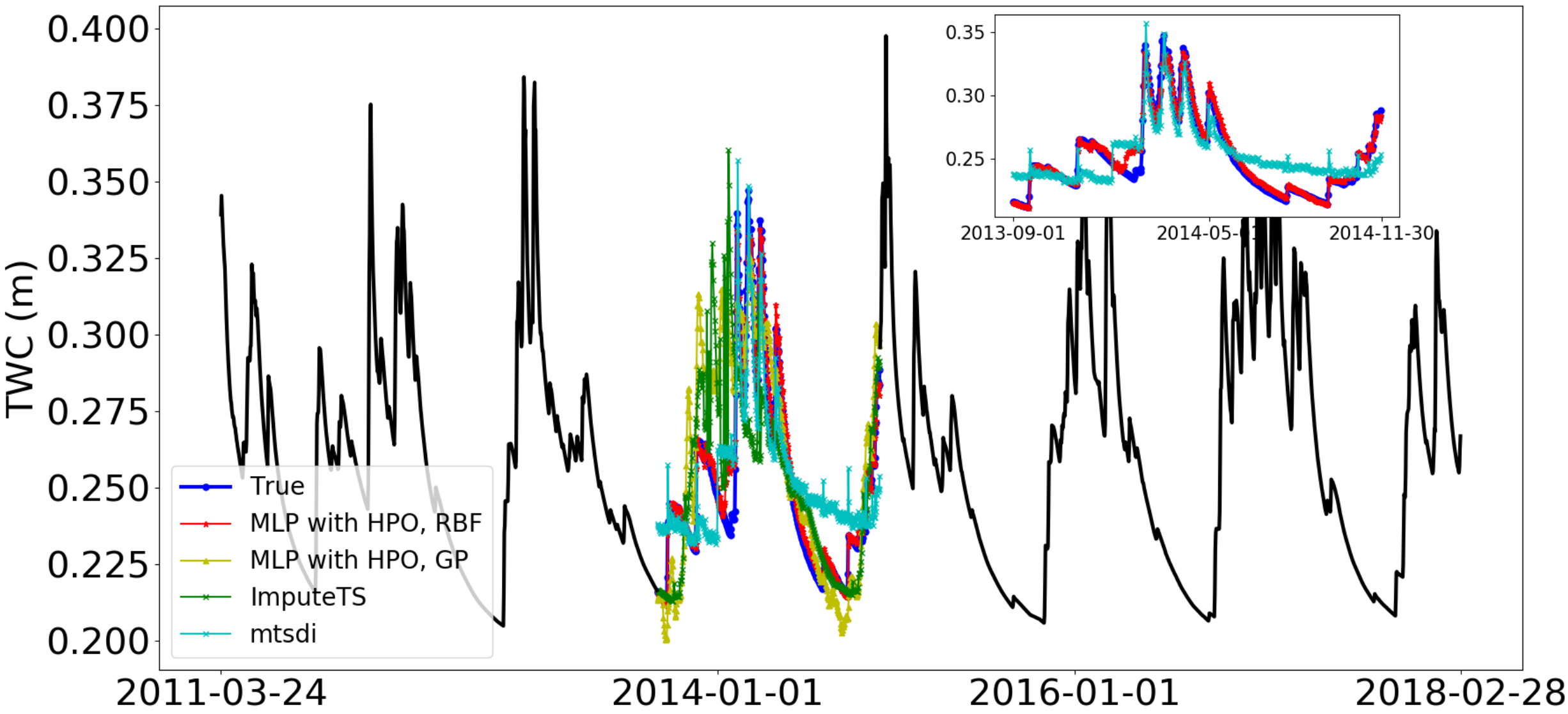}
    \caption{Comparison of gap filling methods for daily TWC with missing values between September 2013 and November 2014. }
    \label{fig:twc_15month}
\end{figure}

Table~\ref{tab:twc_12month} also shows the results  for the case of 15 months of missing data. The MLP-based gap filling method outperformed \textsf{imputeTS} for all instances, but it was again outperformed by \textsf{mtsdi} for most cases (\textsf{mtsdi} had a smaller RMSE). Figure~\ref{fig:twc_15month} shows the results for the gap September 2013-November 2014 (a case in which the MLP based method performed better than the other methods). We can see that the MLP that used the RBF in the HPO led to predictions that almost perfectly matched the true data. \textsf{mtsdi} appeared to be able to capture some of the trends in the data, in particular the large peaks with higher frequency oscillations, but it failed to approximate the ``less rugged" and lower values well. We can also observe from Figure~\ref{fig:twc_15month} how important the choice of hyperparameters is for the MLP. The MLP using the GP during HPO did not perform nearly as well as the MLP that used the RBF during HPO. The main differences between both architectures were in the batch size (50 vs.\ 70), the number of nodes per layer (50 vs.\ 25), and the number of lags (30 vs.\ 345). On the other hand, in Figure~\ref{fig:twc_12month}, the GP-based and RBF-based MLP predictions were approximately equal, and for this case the optimal architectures differed only with respect to the number of nodes per layer (45 vs.\ 25), with the larger architecture (RBF solution) being marginally better than the GP based solution.   Thus, we can recognize the importance of HPO when using DL models and the sensitivity of the MLP's performance with regard to the architecture.

\begin{figure}[htbp]
    \centering
    \includegraphics[width=1.0\textwidth]{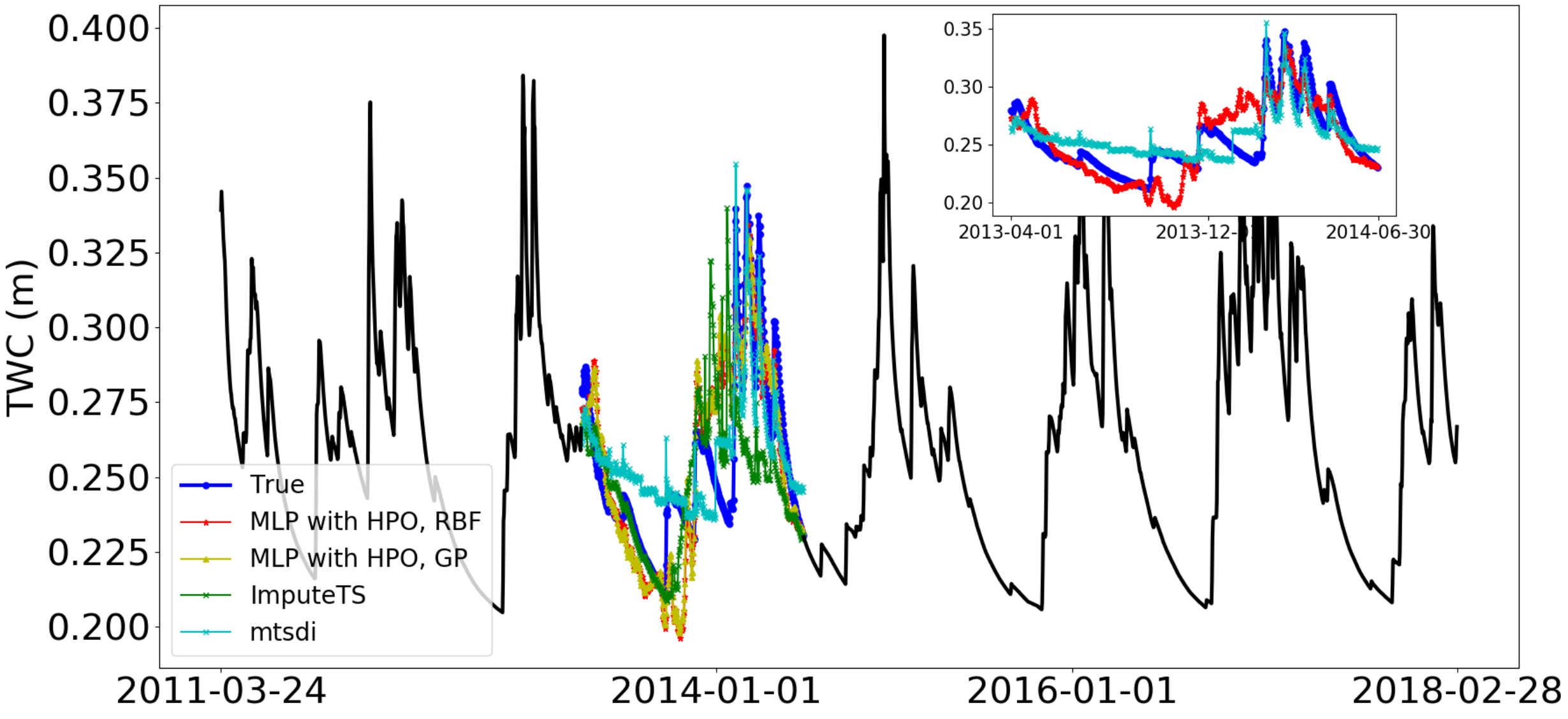}
    \caption{Comparison of gap filling methods for daily TWC with missing values between April 2013 and June 2014. }
    \label{fig:twc_15month1}
\end{figure}

Figure~\ref{fig:twc_15month1} shows the imputation results with 15 months of missing values from April 2013 to June 2014. Similar to Figure~\ref{fig:twc_12month},  \textsf{mtsdi} provided the smallest RMSE but our proposed MLP based methods provided better estimates at the beginning while \textsf{mtsdi} overestimated the values by making an almost constant value prediction. 

\subsubsection{FLUXNET predictions}
\begin{figure}[htbp]
    \centering
    \includegraphics[width=1.0\textwidth]{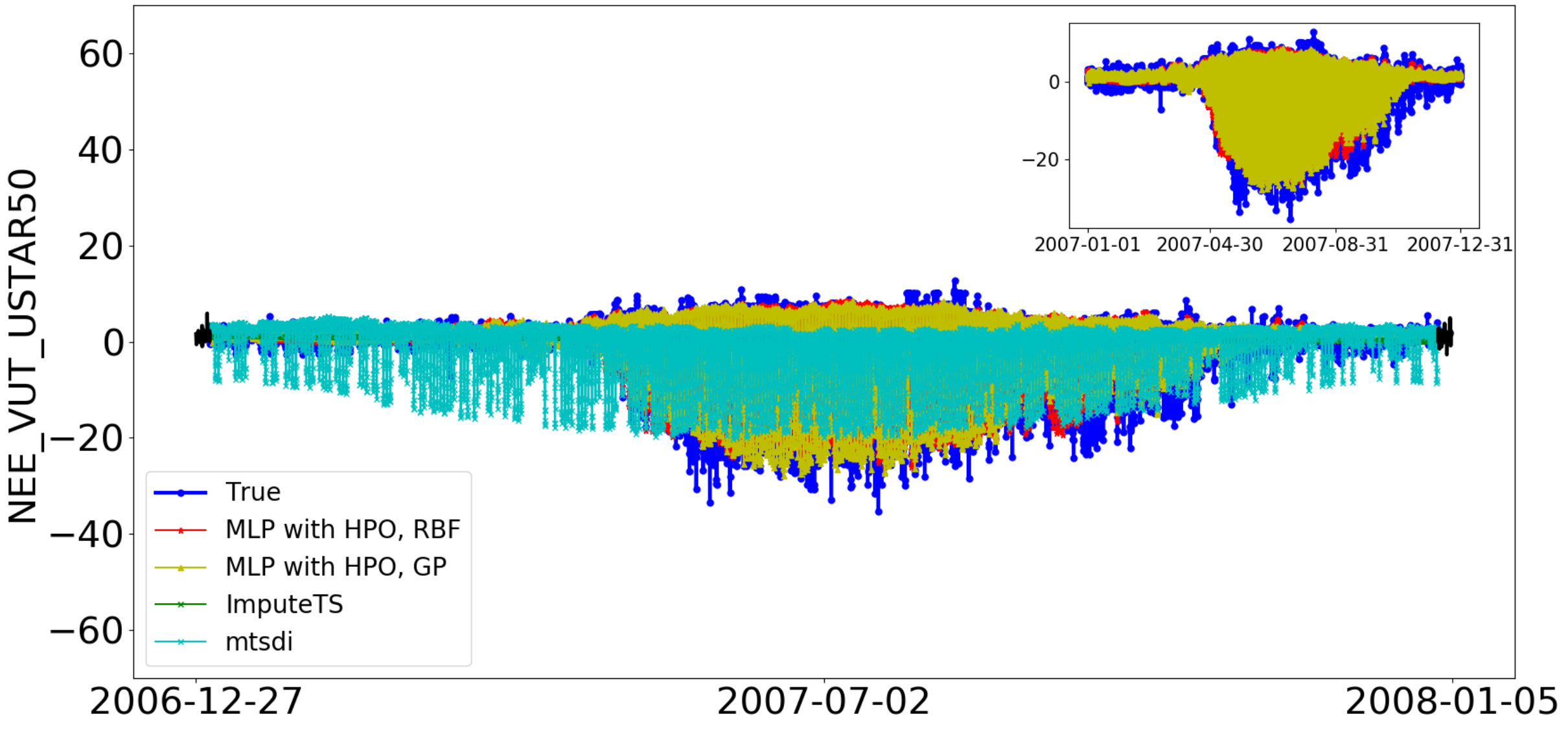}
    \includegraphics[width=1.0\textwidth]{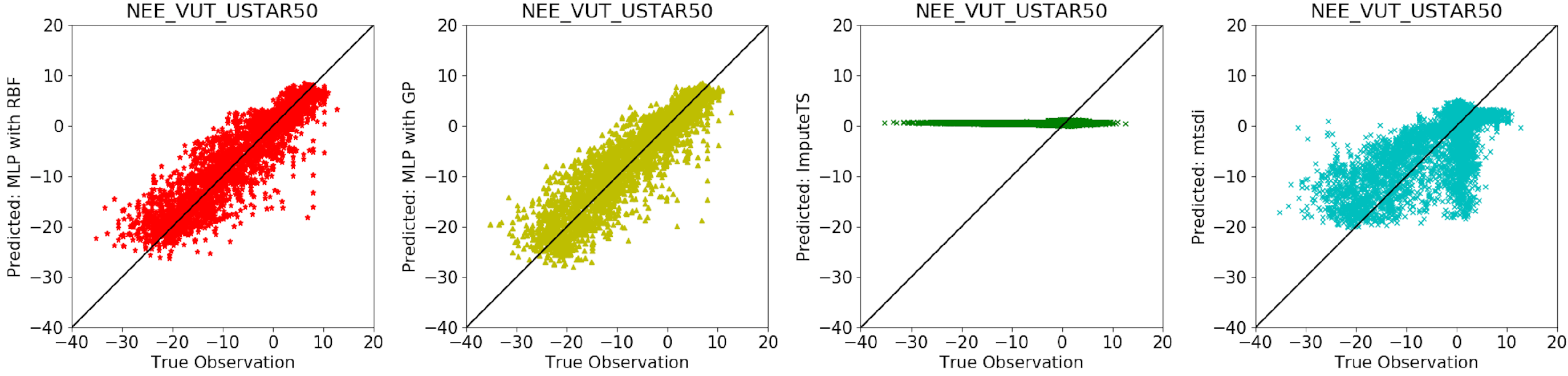}
    \caption{Hourly NEE\_VUT\_USTAR50 (FLUXNET) imputation comparison: missing value range is January 1, 2007 to December 31, 2007. The top panel shows the true and the predicted values. The bottom panels show scatter plots for each method (ideally, prediction and true data are identical and lie on the diagonal)}
    \label{fig:flux3}
\end{figure}

\begin{figure}[htbp]
    \centering
    \includegraphics[width=1.0\textwidth]{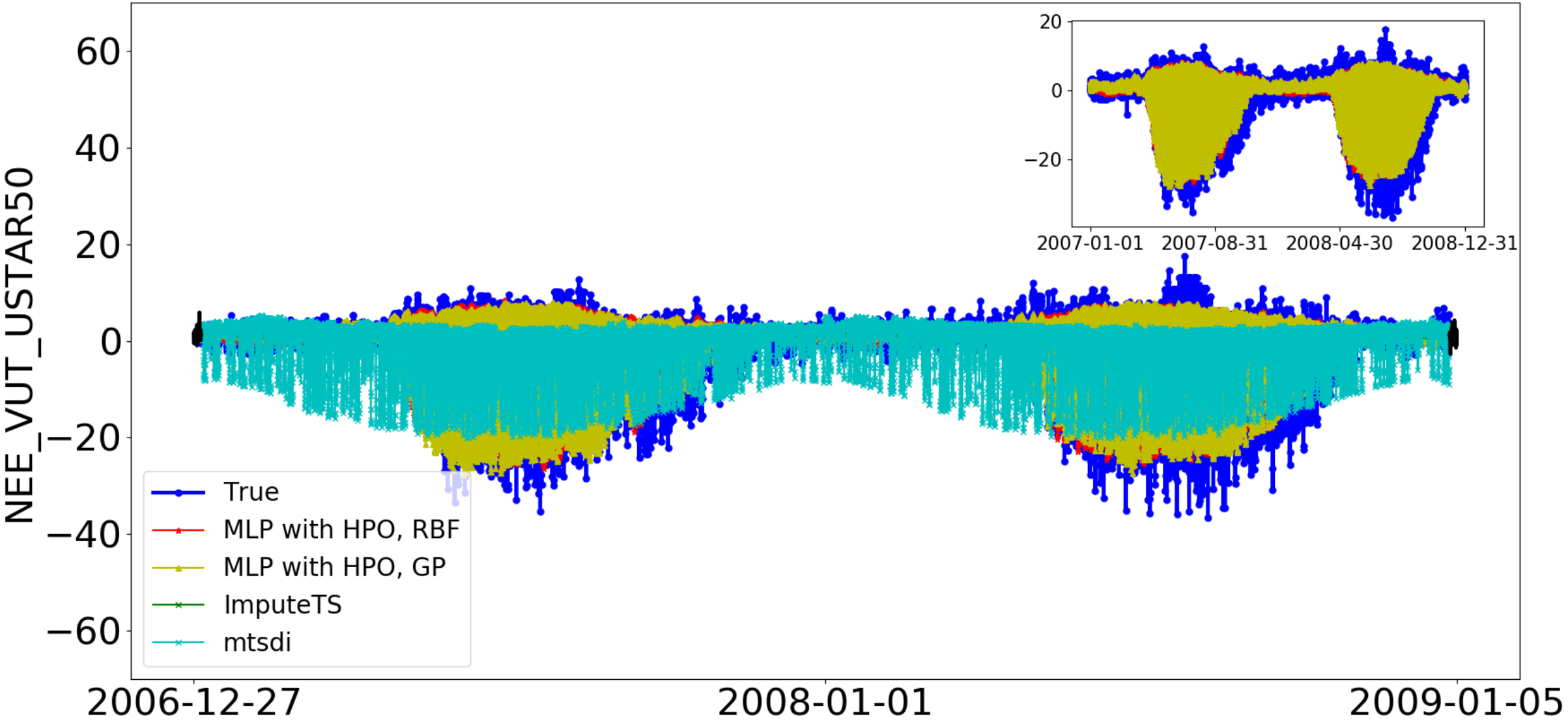}
    \includegraphics[width=1.0\textwidth]{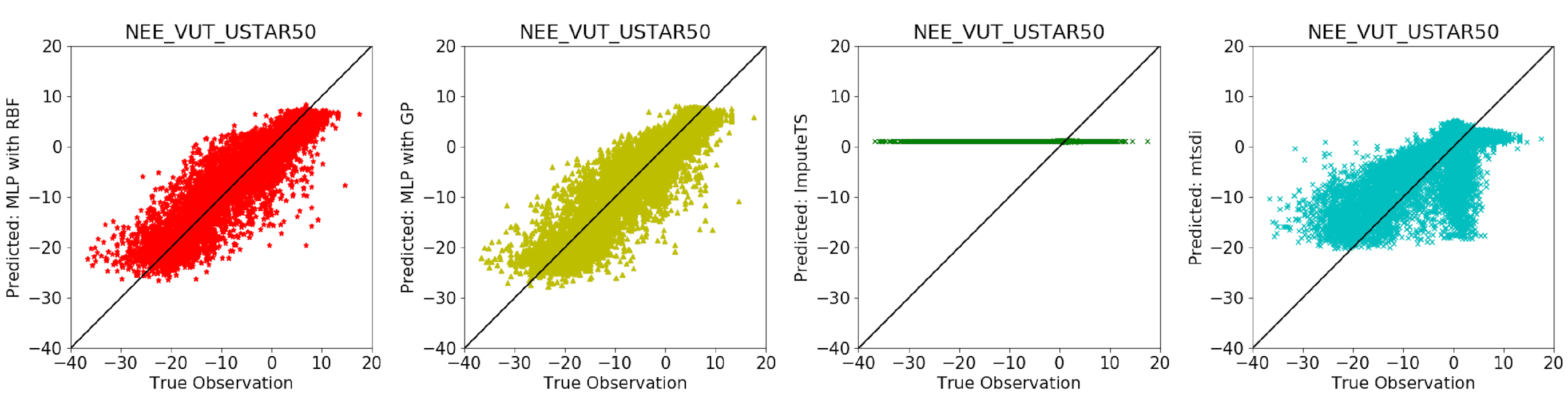}
    \caption{Hourly NEE\_VUT\_USTAR50 imputation comparison: missing value range 2007-01-01--2008-12-31}
    \label{fig:flux4}
\end{figure}

Table \ref{tab:flux} shows the results for the  FLUXNET2015 dataset. Because the number of available observations was large, we stress-tested the performance of the gap filling methods for very large continuous data gaps, using between 2,159 (3 months) and  35,063 (36 months) missing observations. The results show that the MLP-based gap filling method attained better performance (in terms of RMSE) than \textsf{imputeTS}  and \textsf{mtsdi}. The top panel of Figure~\ref{fig:flux3} shows the true and the estimated values of NEE\_VUT\_USTAR50 for the case of 8,759 missing values. The lower panels show scatter plots of the true values versus the predicted values. If the methods were able to exactly repredict the missing values, all points would lie on the diagonal. As can be seen in Figure~\ref{fig:flux3}, \texttt{imputeTS} failed to capture the variability in the data and it simply  connected the last observation (last data point before the gap) with the first observation  after the gap. The MLP-based methods on the other hand made predictions that were close to the true values (points lie close to the diagonal). The point cloud obtained with \textsf{mtsdi} shows a larger variability around the diagonal, and thus for this use case \textsf{mtsdi} made less reliable predictions.   Figure \ref{fig:flux4} plots the results for the case when  17,543 observations are missing. Again, \texttt{imputeTS} failed to capture the trend in the data completely.  \texttt{mtsdi} captured the seasonal trend but it failed to accurately predict the large and small values. Our proposed MLP-based imputation methods  estimated the high and low values significantly better. This is also reflected in the scatter plots which show that the point clouds of  our proposed method lie closer to the diagonal than those of \textsf{imputeTS} and \textsf{mtsdi}. 

Comparing the optimal hyperparameters found by the RBF and the GP methods, respectively, we can see that very different MLP architectures can lead to very similar RMSEs. For example, for the case where 2,159 observations were missing, the difference in RMSEs was only 0.02, but the GP used one extra hidden layer and a significantly larger number of lags. On the other hand, the RBF-based HPO method used more epochs. Similarly, for the case where 17,543 observations were missing, the RBF-based optimizer led to a significantly larger architecture, but it  improved the RMSE by only 0.02. Thus, there is a trade-off between architecture size (and thus training time) and RMSE, and perhaps for large datasets, accepting a slightly larger RMSE in exchange for a simpler and faster to train network is preferable.

\begin{table}[htbp]
  \centering
  \caption{RMSE  for different gap filling methods for   hourly FLUXNET dataset  and hyperparameters ([batch size, \# epochs, \# layers, \# nodes per layer, dropout rate, \# lags]) for both RBF- and GP-based HPO. }
  \resizebox{\textwidth}{!}{\begin{tabular}{ccc|c|c|c|ccc|c}
\cmidrule{1-7}\cmidrule{9-10}    \multicolumn{2}{c}{Missing Range} & \multicolumn{1}{c|}{\multirow{2}[4]{*}{\# obs}} & \multicolumn{4}{c}{RMSE}      &       & \multicolumn{2}{c}{Hyperparameters} \\
\cmidrule{4-7}\cmidrule{9-10}    From  & To    &       & \multicolumn{1}{l|}{RBF} & \multicolumn{1}{l|}{GP} & \multicolumn{1}{l|}{ImputeTS} & \multicolumn{1}{l}{mtsdi} &       & \multicolumn{1}{c|}{RBF} & \multicolumn{1}{c}{GP} \\
\cmidrule{1-7}\cmidrule{9-10}    
     2007-01-01  &  2007-03-31  & 2,159  & 0.98  & \textbf{0.96}  & 1.86  &  4.83 &       & [190, 450,   2,  45,   0,  50]  & [180, 100,   3,  40,   0, 240] \\
     2007-01-01  &  2007-12-31  & 8,759  & 2.72  & \textbf{2.68}  & 8.14  &  5.17 &       & [ 55, 400,   4,  40,   0,  30]  &  [150, 200,   1,  35,   0, 355] \\
     2007-01-01  &  2008-12-31  & 17,543 & \textbf{2.86}  & 2.88  & 8.20  &  5.14 &       & [ 60, 350,   5,  45,   0,  90]  &  [55, 200,   2,  35,   0, 185]\\
     2007-01-01  &  2010-12-31  & 35,063 & \textbf{2.75}  & 2.94  & 8.15  &  5.24 &       & [ 60, 350,   5,  45,   0,  90] &   [70, 500,   1,  25,   0, 345]\\
\cmidrule{1-7}\cmidrule{9-10}    
 \multicolumn{10}{l}{\small\textbf{Bold} numbers indicate the best value for a given problem instance. The column ``\#obs'' indicates the number of missing observations.}\\ \multicolumn{10}{l}{\small``MLP RBF'' and ``MLP GP'' indicate the methods using the MLP with the RBF and GP model during HPO, respectively.}
\end{tabular}}%
  \label{tab:flux}%
\end{table}%



\subsubsection{Discussion}

The  three gap filling case studies  showed that there was no one single method that performed best for all problems and all instances. We found that for the two examples with real observation data (as opposed to simulation-created data), our proposed gap filling method tended to outperform both \textsf{ImputeTS} and \textsf{mtsdi} in terms of RMSE, in particular when the gaps were large. As described in the previous subsections, \textsf{ImputeTS} performed well when the missing data can be approximated by a  somewhat (piecewise) linear function. However, for the FLUXNET2015 data, \textsf{ImputeTS} failed, which may be due to the amount of missing data, or perhaps due to the high frequency oscillations in the data. 
 \textsf{mtsdi} performed better than our method for the simulated total water content data in terms of RMSE, although it did not seem to capture much of the details of the lowest values or the  dynamics in the data. Figures~\ref{fig:twc_12month}  and~\ref{fig:twc_15month1} show that \textsf{mtsdi} seemed to capture the large faster oscillating peaks quite well, but for the slower changes of TWC, it appeared to use a mean value instead of capturing the slower variabilities in the data.  
 
 Although our proposed  imputation approach performed generally well and often outperformed the other methods, its computing time requirement was significantly larger than that of  \textsf{imputeTS} and \textsf{mtsdi}. This is due to the  hyperparameter optimization trying many different sets of hyperparameters and each training is computationally nontrivial.   On the other hand, if the values to be gap filled behave nonlinearly, and if  high prediction accuracy is desired,  we recommend using the DL based approach due to the limitations of the other methods in these cases. In practice, one would not conduct a detailed study as presented here with artificially created gaps, but rather optimize and train the MLP once for the time series to be gap filled.  One way to speed up the DL approach is to either parallelize  the  repetitions of the training  if  compute resources are available or to reduce the number of repetitions. However, the latter option may lead to models that are less reliable in terms of predictive performance. In the FLUXNET test cases, we used only one training  repetition  (instead of five as done with the two other use cases). The MLP based gap filling method always performed  significantly better than the other two methods, indicating that it was not simply a ``lucky" draw. 

Many environmental datasets contain functional relationships between data features.   However, many  time series imputation methods (including the methods \textsf{ImputeTS} and \textsf{mtsdi} discussed in this paper) do not take advantage of this additional information and they use only statistical information contained in the time series data. In other words, they are purely data-driven approaches. 
Unlike these approaches, our proposed DL method can be extended  to include any known physical relationships and  constraints in addition to the statistical information. For example, in the HYDRUS-1D case, we know that there is a mass conservation constraint present between the features and it is possible to include this  constraint, e.g., by adding a penalty term to  the objective function or by including hard constraints during model training.


In our study, we compared the different methods based on their RMSE performance. Although the RMSE is a widely used measure, it does not reflect the individual errors well.  For example, in the daily TWC imputation comparison shown in Figure \ref{fig:twc_12month}, one could argue that the proposed MLP based methods  produce fairly accurate and better predictions than \textsf{mtsdi} for 50\% of the missing values, but they accumulate larger errors for the remaining values. On the other hand, \textsf{mtsdi} does not match most of the data well, but its almost flat prediction accrues overall a lower RMSE.   Thus, it is possible that optimizing for another performance measure may lead to better outcomes for the MLP methods. 

\section{Summary and Conclusions}
\label{sec: con}

In this paper, we proposed a   time series missing value imputation method that uses DL models. Our focus was on time series imputation problems   where a long continuous gap was present  in one variable, and where  all other supporting  variables are fully observed.  Our proposed method uses an MLP whose architecture is tuned with a derivative-free surrogate model based  optimization method. To this end, we modified the hyperparameter optimization approach proposed in~\cite{muller2019surrogate} in order to facilitate the imputation of missing values in time series. 

After training the MLP, we  imputed the missing values sequentially  from the first missing value to the last missing value, basing predictions on previous predictions as we go along. Our sequential   estimation of  missing values allowed us to fill long-term gaps. We chose MLPs as DL models because they have previously  shown good performance for time series data and  because the time needed to optimize the model is significantly shorter than for other types of DL models (see~\cite{muller2019surrogate}).
Generally,  however, any feedforward artificial neural network can be used within our proposed method such as convolutional neural networks \cite{fukushima1980neocognitron, lecun1989backpropagation, lecun1998gradient}. 

We performed  numerical experiments with three different test cases in which we created artificial gaps which allowed us to assess the quality of the gap filled values. We used two test cases with observed values and one test case with simulated data. The numerical results showed that for many problem instances, our proposed approach was able to fill the gaps with  highly accurate values and it was able to capture trends and dynamics in the data better than other data-driven methods such as \textsf{ImputeTS} and \textsf{mtsdi}.   Using  the RBF as surrogate model during HPO  appeared to yield  more stable performance than using the GP. Therefore, we recommend using our proposed approach with the  RBF surrogate during HPO. 

The numerical results also showed that the optimal MLP hyperparameters depended on which data were missing (gap size and location in the time series). Therefore, hyperparameter optimization is essential before using any ML model to impute the missing values. 

The results of this paper can lay the foundation for studying new approaches for  filling long gaps in time series. 
Although the proposed method was able to fill large gaps  with  seasonal trends well, including capturing large and small values, it sometimes failed to correctly estimate the missing values towards  the end of the data gap, indicating a potential for accumulation of errors as we inform predictions with previously predicted values. A ``backward imputation approach'' might be helpful to alleviate this drawback. In this approach, we could  modify the MLP inputs such that we  start imputing data from the end of the gap backwards in time. Combining  both forward and backward imputation may allow us to reap the benefits of both. 

Our study focused on time series that have one large gap.  However, we expect the method to perform reasonably well when we have multiple large gaps in the time series. Since the inputs to the MLP are designed such that they are independent and the HPO is general, we expect that  filling multiple gaps with this method is  straight-forward.
For time series with a mix of large and small gaps, a combination of our DL model based method and other methods that are aimed at filling small gaps can  be used. Incorporation of physics constraints in the DL model predictions has the potential to improve prediction accuracy of environmental data analysis and predictions.  Moreover, if values are missing  in several variables of the multivariate time series, it may be possible to modify the  proposed approach for this scenario. This is a future research topic.

This paper studied the time-series imputation with the MLP which is the most basic feedforward neural network. It would also be interesting to study the proposed time-series imputation approach with other advanced feedforward neural networks such as convolutional neural network \cite{lecun1998gradient} and AR-NET \cite{triebe2019ar}. Recently, a state-of-art time-series forecasting software NeuralProphet \cite{triebe2021neuralprophet} combines the AR-NET and the Prophet software \cite{taylor2018forecasting}.

{\small 
	\section*{Data availability} 
	Three datasets used in this study have been deposited in the GitHub public repository: \hyperlink{https://github.com/JanghoPark-LBL/Dataset_Missing_Value_Imputation_with_Deep_Neural_Networks}{https://github.com/JanghoPark-LBL/Dataset\_Missing\_Value\_Imputation\_with\_Deep\_Neural\_Networks}.
}

{\small 
	\section*{Acknowledgements} 
	This work was supported by Laboratory Directed Research and Development (LDRD) funding from Berkeley Lab, provided by the Director, Office of Science, of the U.S. Department of Energy under Contract No. DE-AC02-05CH11231. This research used resources of the National Energy Research Scientific Computing Center (NERSC), a U.S. Department of Energy Office of Science User Facility operated under Contract No. DE-AC02-05CH11231.
}

\bibliography{sn-bibliography}


\end{document}